\documentclass[11pt]{article}
\usepackage{acl}
\usepackage[T1]{fontenc}
\usepackage[utf8]{inputenc}
\usepackage{times}
\usepackage{latexsym}
\usepackage{microtype}
\usepackage{inconsolata}
\usepackage{soul}
\usepackage{url}
\usepackage{graphicx}
\usepackage{amsmath}
\usepackage{amsthm}
\usepackage{booktabs}
\usepackage{algorithm}
\usepackage{algorithmic}
\usepackage[switch]{lineno}
\usepackage{dblfloatfix}
\usepackage{amssymb}
\usepackage{multirow}
\usepackage{listings}
\usepackage{fvextra}
\DefineVerbatimEnvironment{promptblock}{Verbatim}{
    fontsize=\small,
    breaklines=true,
    breaksymbolleft={},
    breaksymbolright={},
    frame=lines,
    commandchars=\\\{\},
}
\usepackage{xcolor}
\usepackage{array}
\usepackage{makecell}
\usepackage{longtable}
\usepackage{tabularx}
\usepackage{siunitx}
\usepackage{threeparttable}
\usepackage{needspace}
\usepackage{ragged2e}
\usepackage{enumitem}

\title{Real-Time Deadlines Reveal Fragile Temporal Adaptation in LLM Strategic Dialogues}

\author{
Neil Sehgal$^1$ \and Sharath Chandra Guntuku$^1$ \and Lyle Ungar$^1$ \\
$^1$University of Pennsylvania \\
\texttt{\{nsehgal, sharathg, ungar\}@upenn.edu}
}

\begin{document}

\maketitle
\begin{abstract}
Large Language Models (LLMs) generate text token-by-token in discrete time, yet real-world communication, from therapy sessions to business negotiations, critically depends on continuous time constraints. We use simulated negotiations between paired agents under strict deadlines to study adaptation to real-time pressure. Agents either receive only the initial deadline or explicit remaining-time updates at each turn. Remaining-time feedback raises deal closure from 4\% to 32\% for \texttt{GPT-5.1-chat-latest} and increases offer acceptance more than sixfold. The same model achieves near-perfect closure under turn-based limits, showing that poor wall-clock performance is not simply due to insufficient negotiation competence. Across additional interface conditions, qualitative urgency cues can outperform numeric countdowns, repeated deadline reminders do not consistently reproduce their benefits, and directed time tracking can help or hurt depending on the model. Across additional negotiation scenarios and model configurations, we find real-time temporal adaptation is fragile, model-dependent, and sensitive to how temporal constraints are presented. \footnote{Code available at https://github.com/sehgal-neil/llm-temporal-awareness}


\end{abstract}

\section{Introduction}
Large Language Models (LLMs) are increasingly deployed as interactive agents, serving as copilots in productivity tools, customer-service representatives, tutoring assistants, and decision-support systems. In these settings, performance depends not only on what agents say but how their behavior adapts over time. Human interaction unfolds in continuous time: deadlines, delays, and interruptions shape communication and strategy. Yet, standard LLM evaluation abstracts away from continuous time, modeling interaction as discrete turns unconstrained by real deadlines.

Humans adapt as time progresses. Patients raise concerns as appointments end \citep{nielsen2012patient,white1994oh,white1997wrapping}; negotiators concede near deadlines \citep{aer_deadline_effect}; auction bidders place last-second bids \citep{backus2015sniping}. These patterns reflect sensitivity to time’s passage.

Consider a negotiation assistant that has 30 seconds remaining before a deadline expires. A decent offer is on the table. Should it accept or hold out for marginal improvements? Recent work suggests LLMs struggle with temporal tracking \citep{wang2025discretemindscontinuousworld,cheng2025temporalblindnessmultiturnllm}. If the model fails to recognize time is nearly up, it may hold out and lose the deal. This may happen even when strategic reasoning is otherwise adequate, because the model does not reliably infer and act on the remaining time.

This limitation has practical consequences. A negotiation assistant may hold out and lose deals. A triage assistant may misallocate attention as time runs out. An auction agent may bid rationally yet still lose by acting too early. Ignoring real time can produce systematic errors even when time-independent reasoning is otherwise strong.

We refer to the capacity required for effective performance in such settings as \emph{temporal awareness}. We define temporal awareness as the ability to (1) represent how much time has elapsed and remains, (2) anticipate how others’ behavior changes as time passes, and (3) condition one's own strategy on the current temporal state. Temporal awareness is therefore a core component of intelligent strategic interaction in the wild, enabling agents to adapt as opportunities narrow and constraints bind.

These observations raise a central question: \textbf{Can LLM agents internalize time pressure during multi-turn interaction and adapt their strategic behavior accordingly?}

\begin{figure*}[!t]
    \centering
    \includegraphics[width=.8\linewidth]{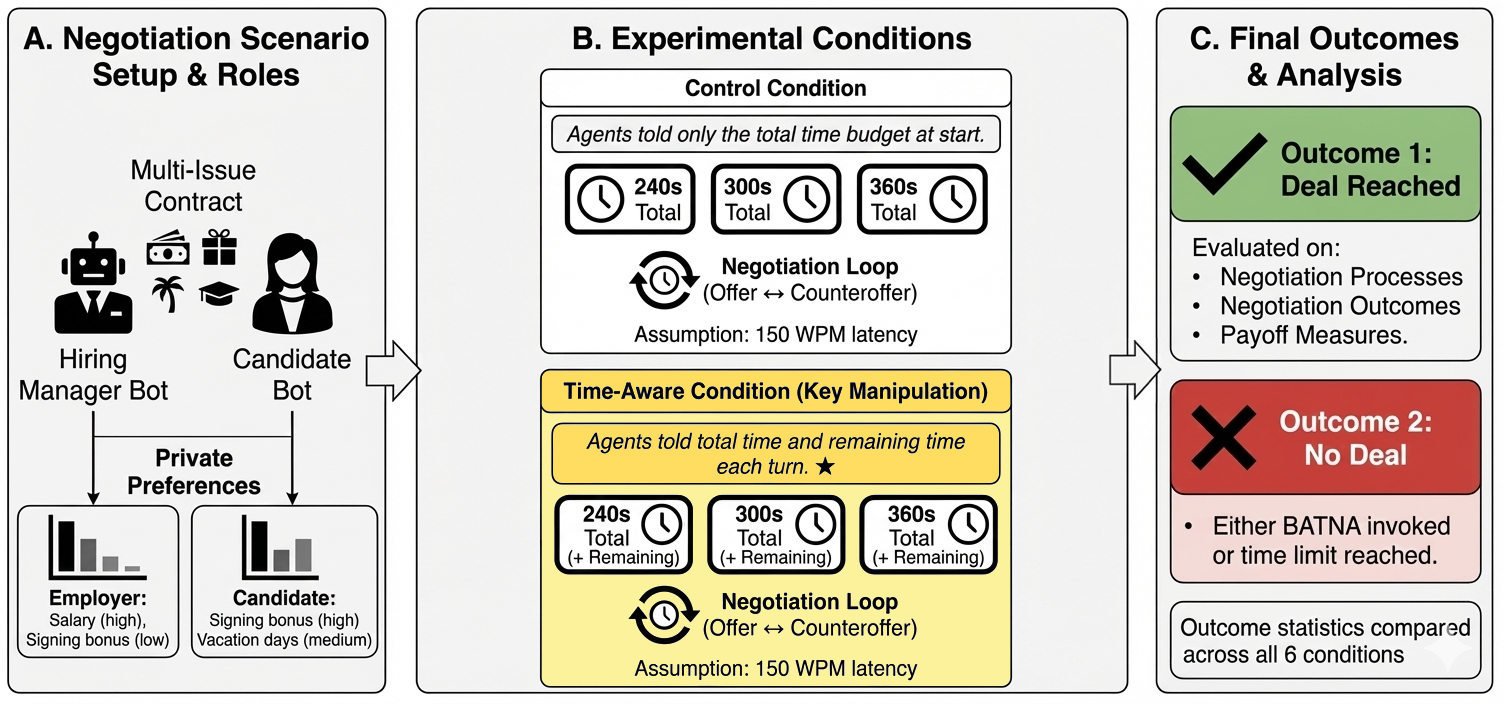}
    \caption{\textbf{Experimental Design and Study Flow.} Two LLM agents negotiate over a multi-issue hiring package under varying levels of time pressure (240, 300, or 360s), assuming a fixed generation latency of 150 WPM (words per minute). In the \textsc{Control} Condition, agents are informed only of the total time available. In the \textsc{Time-Aware} Condition, agents receive the total time limit and explicit updates on remaining time at each turn. Each negotiation ends when agents reach agreement, take their outside option (BATNA), or exhaust their allotted time or turns.}
    \label{fig:spirit}
\end{figure*}

To answer this question, we introduce a controlled negotiation paradigm in which paired LLM agents engage in multi-issue hiring negotiations under strict real-time deadlines (Figure~\ref{fig:spirit}). We vary deadline length and manipulate temporal information. In a \textsc{Control} condition, agents are informed only of the total time available. In a \textsc{Time-Aware} condition, they additionally receive explicit updates on the remaining time at each turn. This design holds incentives, issues, and turn structure constant while varying agents’ access to temporal state. As a result, differences in outcomes measure the behavioral effect of how temporal information is surfaced.

Across models, negotiation settings, and deadlines, we find a consistent dissociation. Agents reach near-perfect deal closure under turn limits, while performance under wall-clock limits often improves substantially when remaining time is explicitly surfaced. Temporal feedback raises offer acceptance odds more than sixfold and sharply improves deal closure rates without degrading offer quality or dialogue length. These findings show that effective adaptation to continuous-time constraints does not emerge reliably under a one-time deadline specification.

In summary, this work:

\begin{enumerate}[nosep,leftmargin=*]
\item \textbf{Formulates temporal awareness} as an underexplored dimension of LLM-agent evaluation, distinct from turn-based reasoning.
\item \textbf{Introduces a controlled, multi-issue negotiation framework} isolating real-time deadlines from strategic complexity, enabling direct tests of LLM agents’ sensitivity to continuous time.
\item \textbf{Demonstrates empirically} across models and scenarios that explicit temporal feedback alters negotiation outcomes, suggesting current LLMs often fail to internalize elapsed time.
\end{enumerate}

\section{Related Work}

\textbf{Temporal reasoning in LLMs.} 
Prior work on temporal reasoning has mainly focused on static question answering. Benchmarks like MCTACO and TimeDial \citep{zhou-etal-2019-going,qin-etal-2021-timedial} show LLMs can handle basic temporal QA but struggle with more complicated temporal dependencies. However, these assess offline comprehension, not real-time tracking during interactions.

\textbf{LLMs as strategic agents.} Recent work has explored LLM-driven bargaining and cooperation, including buyer–seller price negotiations and multi-agent coordination benchmarks. These studies reveal LLMs can participate in structured strategic interactions. However, performance can depend on role-specific prompting or fine-tuning \citep{kim2024leveraging,fossey2025argument,bianchi2024well,abdelnabi2023llm,kwon2024llms}. Multi-agent frameworks similarly demonstrate emerging coordination, but rely on fixed turn budgets \citep{multiagentbench}. These settings rarely involve continuous deadlines. This leaves  untested whether models adapt strategically as real-time pressure increases.

\textbf{Human negotiation under time pressure.} Human behavioral research consistently finds  deadlines shape concession patterns and agreement timing. Research finds concessions increase and agreements cluster near deadlines, and individualistic parties produce non-agreements and poor outcomes \citep{stuhlmacher1998impact,aer_deadline_effect,carnevale1986time}. These findings motivate evaluating whether LLM agents exhibit similar temporal adaptation.

\textbf{Temporal awareness and elapsed-time adaptation in agents.} A nascent line of work investigates whether LLMs adjust their behavior as real or simulated time elapses during an interaction.
One recent study finds models shorten responses when users explicitly express urgency. However, these effects vary by task difficulty and model \citep{wang2025discretemindscontinuousworld}. Another study documents temporal blindness in tool-use scenarios, where agents fail to recognize when cached information is stale \citep{cheng2025temporalblindnessmultiturnllm}. Together, this literature suggests current models can react to overt temporal signals but do not robustly use elapsed-time information to guide downstream behavior. What remains unclear is whether similar limitations arise in multi-turn strategic interactions, where optimal concessions, pacing, and acceptance decisions should change continuously as real deadlines approach. We therefore directly test whether LLM agents can track and respond to real time in open-ended negotiation, isolating temporal awareness from strategic competence by holding incentives and dialogue structure fixed while varying whether remaining time is explicitly surfaced.

\section{Experimental Negotiation Framework}
\label{sec:exp_framework}
We study whether contemporary LLMs can represent and act on the passage of \emph{continuous} time during multi-turn strategic interaction. We use negotiation as a testbed because success depends not only on static strategic reasoning but also on adapting behavior as deadlines approach. Under time pressure, a capable agent should adjust pacing, concessions, exploration of offers, and acceptance decisions. Throughout the paper we use the term \emph{agent} to refer to an LLM instantiated within the negotiation environment with a fixed system prompt, dialogue history, and structured action interface. The underlying model generates each action, while the agent refers to the model interacting with the environment through this protocol. Agents are not given access to an external clock or timing tool; any time-sensitive behavior in the \textsc{Control} condition must therefore arise from reasoning about the initial time limit and the progression of dialogue turns.

This challenge is particularly visible in multi-issue bargaining. In simple zero-sum price negotiations, time pressure mainly affects concession size along a single bargaining dimension. In multi-issue negotiation, however, time pressure also limits exploration of the agreement space and the discovery of mutually beneficial cross-issue tradeoffs. This makes multi-issue negotiation a stronger setting for testing temporal adaptation: models must preserve strategic competence while opportunities to explore and coordinate shrink over time.

To test this, we construct a controlled bilateral negotiation task in which two agents bargain over a multi-issue contract under strict deadlines. Our key design choice is to manipulate only agents' access to temporal state. In the \textsc{Control} condition, agents are told the total time limit once at the beginning of the episode but receive no further temporal feedback. In the \textsc{Time-Aware} condition, agents additionally receive explicit updates about remaining time at each turn. All other components of the environment are held fixed, including roles, payoff structure, action space, and interaction protocol.

\subsection{Task and negotiation environment}
\label{subsec:env}

Each episode instantiates a bilateral, multi-issue negotiation adapted from existing, widely used case studies \citep{new-recruit,rubbermind}, with two agents playing the roles of hiring manager and job candidate. The agents negotiate over a fixed set of $m$ contract issues (e.g.\ salary, signing bonus, vacation days, start date), where each issue $i \in \{1,\dots,m\}$ takes a value in a finite discrete domain $\mathcal{D}_i$. A complete contract outcome comprises a bundle
\vspace{-8pt}
\[
x = (x_1,\dots,x_m) \in \mathcal{D}_1 \times \cdots \times \mathcal{D}_m.
\]
This design choice also distinguishes our task from a common evaluation pattern in LLM-agent bargaining work that emphasizes buyer--seller price negotiation, which is often effectively modeled as a single distributive issue (i.e.\ zero-sum) \citep{xia2024measuring,bianchi2024well}.

\paragraph{Private payoffs and general-sum structure.}
Each agent $a \in \{\text{manager}, \text{candidate}\}$ receives a private payoff table that assigns values to the options of each issue. We model utilities as additive across issues:
\vspace{-8pt}
\[
U_a(x) \;=\; \sum_{i=1}^{m} u_a^i(x_i).
\]
Under this specification, there are no cross-issue interaction terms; strategic complexity arises from private information and the need to identify mutually beneficial cross-issue tradeoffs (logrolling) \citep{priya2025genteel}. Although some issues are distributive (i.e. agents' preferences are opposed), the overall negotiation is general-sum due to the presence of compatible (identical preferences) and integrative (mutually beneficial trades exist) structure in the payoff tables. Intuitively, this means there exist agreements that are better for both parties than other agreements (i.e. 'win-win' trades). Parties must identify these through strategic communication, which becomes harder under time pressure.



\paragraph{Action space and termination.}
The dialogue alternates between agents. At each turn, an agent produces (i) a natural-language message and (ii) a structured proposal for a contract $x$. The other agent may accept the latest proposal, counter with a new proposal, or invoke an outside option (Best Alternative To a Negotiated Agreement; BATNA). Each episode terminates when: (1) a proposal is accepted; (2) an agent invokes its BATNA; or (3) the interaction exhausts its allotted budget (time or turns, depending on condition).

\subsection{Temporal state and conditions}
\label{subsec:temporal}

We model time as a continuous environment variable that decreases as the negotiation progresses. Each episode begins with a wall-clock budget
$T \in \{240, 300, 360\}$ seconds. Let $\tau_t$ denote the remaining time (in seconds) immediately before turn $t$. Our key manipulation is whether $\tau_t$ is included in the agents' observation.

\paragraph{Time-Limit-Only (Control).}
Agents are informed of the total budget $T$ once at the beginning of the episode, but do not observe $\tau_t$ thereafter. Any time-sensitive behavior therefore requires internal tracking of elapsed time across turns.

\paragraph{Time-Aware (Remaining-Time Feedback).}
Agents are informed of the same total budget $T$ at the beginning and additionally receive the current remaining time $\tau_t$ at \emph{every} turn via an explicit textual prefix on the other agent's message (e.g. ``\texttt{137 seconds left}'').

This design creates a clean informational contrast: if agents can robustly track elapsed time internally, providing $\tau_t$ should minimally impact outcomes. Systematic improvements indicate that externally supplied temporal state can affect behavior. However, this contrast alone does not distinguish failures of internal tracking from alternative explanations such as deadline salience or policy activation.

\paragraph{Follow-up interface ablations.}
We additionally evaluate two follow-up conditions designed to probe two alternative explanations for the Time-Aware advantage: repeated deadline salience and the absence of directed temporal self-tracking. In the \textbf{Total-Budget Reminder} condition, agents receive a per-turn reminder of the original total budget but no remaining-time information. In the \textbf{CoT Time-Tracking} condition, agents are explicitly instructed to estimate elapsed and remaining time before acting, without access to an external clock. We restrict these ablations to \texttt{GPT-4.1}, \texttt{GPT-5.1-chat-latest}, and \texttt{GPT-5-mini}, for which follow-up comparisons could be interpreted without large floor effects or reasoning-throughput confounds.\footnote{Among the remaining configurations, reasoning-disabled \texttt{Qwen3-8B} was the only additional model with substantial original closure and without a reasoning-throughput confound. We attempted the follow-up ablations for this configuration, but the original OpenRouter runs had been served by Fireworks, which no longer exposed the checkpoint. A replacement serving configuration did not reproduce the original Control or Time-Aware baseline, so we excluded those non-equivalent runs.} We run $N=100$ negotiations per model, deadline, and treatment. Each ablation was collected in a separate batch together with contemporaneous \textsc{Control} and \textsc{Time-Aware} baselines. Because these ablations occurred several months after the original experiments (May 2026), and absolute baseline rates varied across batches, we compare treatments only against the contemporaneous baselines from the same batch and do not compare absolute rates across periods.

\subsection{Interaction protocol and simulation}
\label{subsec:protocol}

\paragraph{Structured action format.}
To standardize behavior across models and enable reliable parsing, agents emit a JSON-formatted action each turn containing: (i) a natural-language message, (ii) a structured offer over the issue fields (or a null offer), and (iii) binary indicators for acceptance and BATNA invocation. When either acceptance or BATNA is \texttt{true}, the episode terminates. Agents receive the full dialogue history each turn (system prompt and prior utterances), and transcripts remain short (mean (SD): 471 (192) words), so results are not driven by context-window overflow, history truncation, or summarization. All negotiations use the Kani multi-agent LLM framework \citep{zhu2023kani}. Full prompts and templates appear in Appendix~\ref{app:prompts}. 

\paragraph{Real-time simulation and message latency.}
To approximate human dialogue pacing constraints, the  engine applies a speech-latency model: each utterance incurs a delay proportional to word count, calibrated at 150 words per minute \citep{tauroza1990speech}, deducted from the remaining time budget. Latency is computed only from the natural-language message, excluding JSON fields (e.g. structured offers) and reasoning tokens. This latency serves two purposes: (1) approximates realistic dialogue pacing, and (2) creates a strategic tradeoff between elaboration and urgency under deadlines. The rule is disclosed in both agents' system prompts. Removing latency leaves the qualitative pattern unchanged (Section~\ref{subsec:latency}).

\paragraph{Scenarios and counterbalancing.}
We evaluate two negotiation scenarios drawn from widely used negotiation case studies, structured role-play simulations commonly used in management and negotiation training. The primary scenario adapts the ``New Recruit'' case, and we additionally test the ``Rubbermind'' case to assess generalization across issue sets and valuation structures \citep{new-recruit,rubbermind}. For each scenario, treatment, and time limit, we run 100 simulated negotiations and counterbalance which role initiates.

\paragraph{Recorded outcomes.}
We record negotiation-level outcomes (e.g. if deals are reached before expiration; final joint payoff) and offer-level events (e.g. if a given proposal is accepted; the recipient's payoff under the proposal). This enables analyses of overall deal closure and acceptance dynamics under time pressure.

\section{Results}

We begin by analyzing \texttt{GPT-5.1-chat-latest}, which exhibits among the highest average deal-closure rates in our experiments, and then assess robustness to turn-based constraints, negotiation scenario, latency assumptions, model families (including open and closed, reasoning and non-reasoning models), and interface ablations. \texttt{GPT-5.1-chat-latest} is a conversation-optimized model that powered ChatGPT at the time of our experiments, making it particularly representative of deployed interactive agents and well-suited for diagnosing failures in real-time, multi-turn interaction.

\subsection{Explicit Temporal Feedback Improves Performance Under Time Pressure}

We first ask how explicit time awareness affects negotiation outcomes under real-time deadlines.  Two \texttt{GPT-5.1-chat-latest} agents negotiate a multi-issue hiring contract under three time limits (240, 300, and 360 seconds). In the \textsc{Control} condition, both agents are told only the total time available at the start. In the \textsc{Time-Aware} condition, they are told the same total time and also receive explicit feedback on the remaining time at each turn. The underlying payoff structure and outside options are identical across conditions, and we run 100 negotiations per deadline per condition (N = 300 per condition; 600 total). This experiment tests our central hypothesis: if LLMs can internally track time, explicit temporal feedback should have minimal impact on outcomes.

\begin{figure}
    \centering
    \includegraphics[width=1\linewidth]{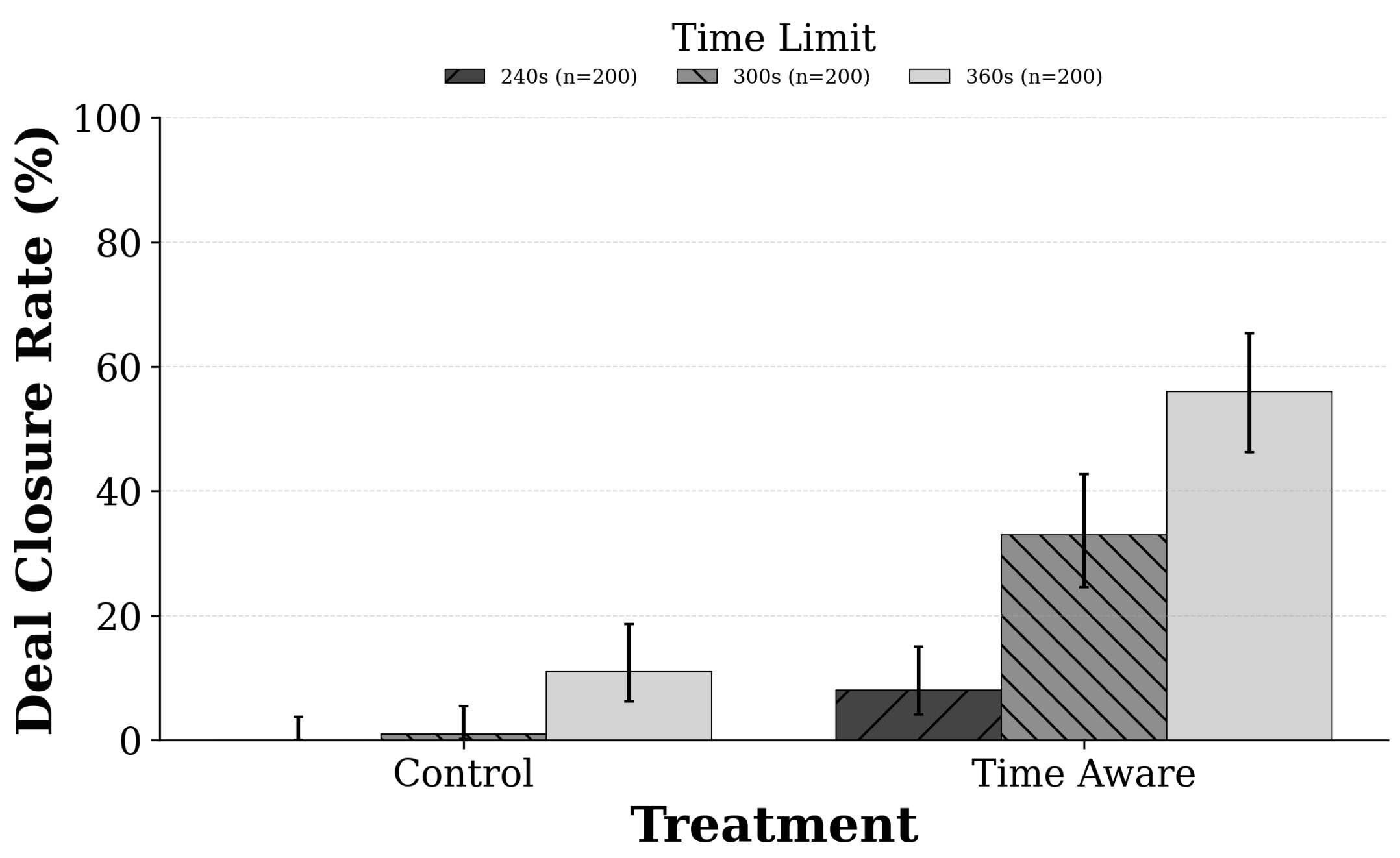}
    \caption{\textbf{Deal Closure by Treatment and Time Limit.} Deal closure rates for \texttt{GPT-5.1-chat-latest} across conditions and time limits (N=100 per condition). \textsc{Time-Aware} agents, who receive explicit temporal feedback each turn, achieve higher closure rates than \textsc{Control} agents at every deadline. Error bars represent 95\% Wilson confidence intervals.}

    \label{fig:fig2}
\end{figure}

Figure~\ref{fig:fig2} shows making agents time-aware increases deal closure rates at every deadline. Aggregated across time limits, negotiation-level closure rates increase from 0.04 (12/300) in \textsc{Control} to 0.32 (97/300) in the \textsc{Time-Aware} condition, a 708\% relative improvement. These gains are not solely explained by conversation length or reasoning effort: the mean words per negotiation and mean reasoning tokens per utterance are similar across conditions ($p>0.05$ for both) (Table \ref{tab:table1}).

\begin{table}[t]
\centering
\footnotesize
\setlength{\tabcolsep}{3.5pt}
\renewcommand{\arraystretch}{1.08}

\begin{tabularx}{\columnwidth}{@{}>{\raggedright\arraybackslash}Xccc@{}}
\toprule
Metric & Control & Time-aware & $\Delta$ \\
\midrule
\multicolumn{4}{@{}l@{}}{\textbf{Process}} \\
No. utterances\textsuperscript{***}        & 6.7 (1.3)   & 7.3 (1.5)   & +7.9\% \\
Words / negotiation                         & 627 (116)   & 634 (115)   & +1.1\% \\
Words / utterance\textsuperscript{***}     & 93 (23)     & 87 (24)     & -6.3\% \\
Reasoning tokens / utterance                & 65 (23)     & 67 (23)     & +3.1\% \\
\addlinespace[2pt]

\multicolumn{4}{@{}l@{}}{\textbf{Outcomes}} \\
Deal rate\textsuperscript{***}             & 0.04        & 0.32        & +708\% \\
Acceptance rate\textsuperscript{***}       & 0.6         & 4.5         & +649\% \\
\addlinespace[2pt]

\multicolumn{4}{@{}l@{}}{\textbf{Payoff}} \\
First-offer joint payoff                    & 20088 (1047)& 20046 (1180)& -0.2\% \\
Final joint payoff                          & 19476 (1309)& 19454 (2379)& -0.1\% \\
\bottomrule
\end{tabularx}

\caption{\textbf{Negotiation process and outcome metrics for Control and Time-Aware conditions (\texttt{GPT-5.1-chat-latest}).} N=300 per condition. $\Delta$ is relative to Control. Payoffs in scenario utility points. *** $p<0.001$.}
\label{tab:table1}
\vspace{-8pt}
\end{table}

Offer content also changes minimally. First-offer and final joint payoffs are
nearly identical across conditions ($p > 0.05$) (Table \ref{tab:table1}). Yet offers in the \textsc{Time-Aware} condition are accepted
much more often. To quantify this, we model the probability an offer is accepted using a logistic regression with indicators for deadline length, a \textsc{Time-Aware} treatment indicator, and the receiving agent’s payoff (Table~\ref{tab:appendix_log}).

Controlling for deadline and payoff, offers in the \textsc{Time-Aware} condition are over six
times more likely to be accepted (OR = 6.38, $p < .001$). Longer
deadlines also increase acceptance probabilities: relative to 240\,s,
a 300\,s limit raises the odds of acceptance by a factor of 3.42 ($p < .001$), and a
360\,s limit by a factor of 4.97 ($p < .001$). As expected, more
generous offers are also more likely to be accepted (OR = 2.01 per
1{,}000 utility points, $p < .001$), indicating the agents respond sensibly to
payoff differences.


These results show that supplying remaining-time information can substantially improve negotiation behavior without materially changing offer quality or overall dialogue length. The \textsc{Control–Time-Aware} contrast alone does not distinguish failures of internal time estimation from deadline salience or policy activation; Sections 4.3 and 4.4 examine these alternatives.

\subsection{Turn Limits Reveal High Strategic Competence}
\label{subsec:turn_limits}
To assess whether poor performance under time pressure could instead reflect inadequate strategic competence, we compare time-limited negotiations to turn-limited controls. In the turn-limited setting, agents must reach agreement within 5, 6, 7, 8, or 9  turns. The payoff structure, outside options, and protocol are unchanged.

Under turn limits, \texttt{GPT-5.1-chat-latest} agents achieve near-perfect deal closure rates across all budgets. With 5 total utterances, 99\% of negotiations reach agreement; with 6 utterances, 98\%; with 7 utterances, 99\%; with 8 utterances 100\%; with 9 utterances, 99\%.  Moreover, in all 5 negotiations that ended without agreement, one agent deliberately invoked its BATNA before exhausting turns.


This dissociation shows that poor performance under wall-clock deadlines is not simply caused by an inability to negotiate under constrained interaction. Agents can use discrete turn budgets to structure concessions and reach agreement. However, turn limits remove the need to estimate elapsed time, eliminate variable utterance-cost tradeoffs, and change the policy space. We therefore treat this experiment as a diagnostic contrast rather than a matched counterfactual. The remaining wall-clock gap may reflect temporal-state estimation, deadline salience, policy activation, or other interface-dependent factors.

\subsection{LLMs Respond to Decision-Local Urgency Cues}
\label{subsec:urgency_ablation}

The strong turn-limit results suggest agents have ample strategic competence when constraints are expressed in token-aligned units. A remaining question is why agents fail under real-time deadlines when the time limit is only given once. One possibility is models are broadly ``deadline insensitive'' in negotiation. Another possibility is that they can respond to local urgency cues, but fail to generate those adjustments from an internal estimate of elapsed time when temporal feedback is absent.

To distinguish these explanations, we introduce an \textsc{Urgency} ablation that mirrors the \textsc{Time-Aware} interface in format (a short per-turn prefix) but removes numeric remaining-time state. Concretely, in the \textsc{Urgency} condition, each agent receives a brief qualitative reminder at every turn ``\texttt{(Deadline approaching--act with urgency.)}'', while the \textsc{Time-Aware} condition receives the numeric countdown (e.g. ``\texttt{(137 seconds left)}'') and the \textsc{Control} condition receives only a one-time statement of the total time budget $T$ at the start. All payoff tables, BATNAs, protocol, and termination rules are unchanged.

\begin{figure}
    \centering
    \includegraphics[width=1\linewidth]{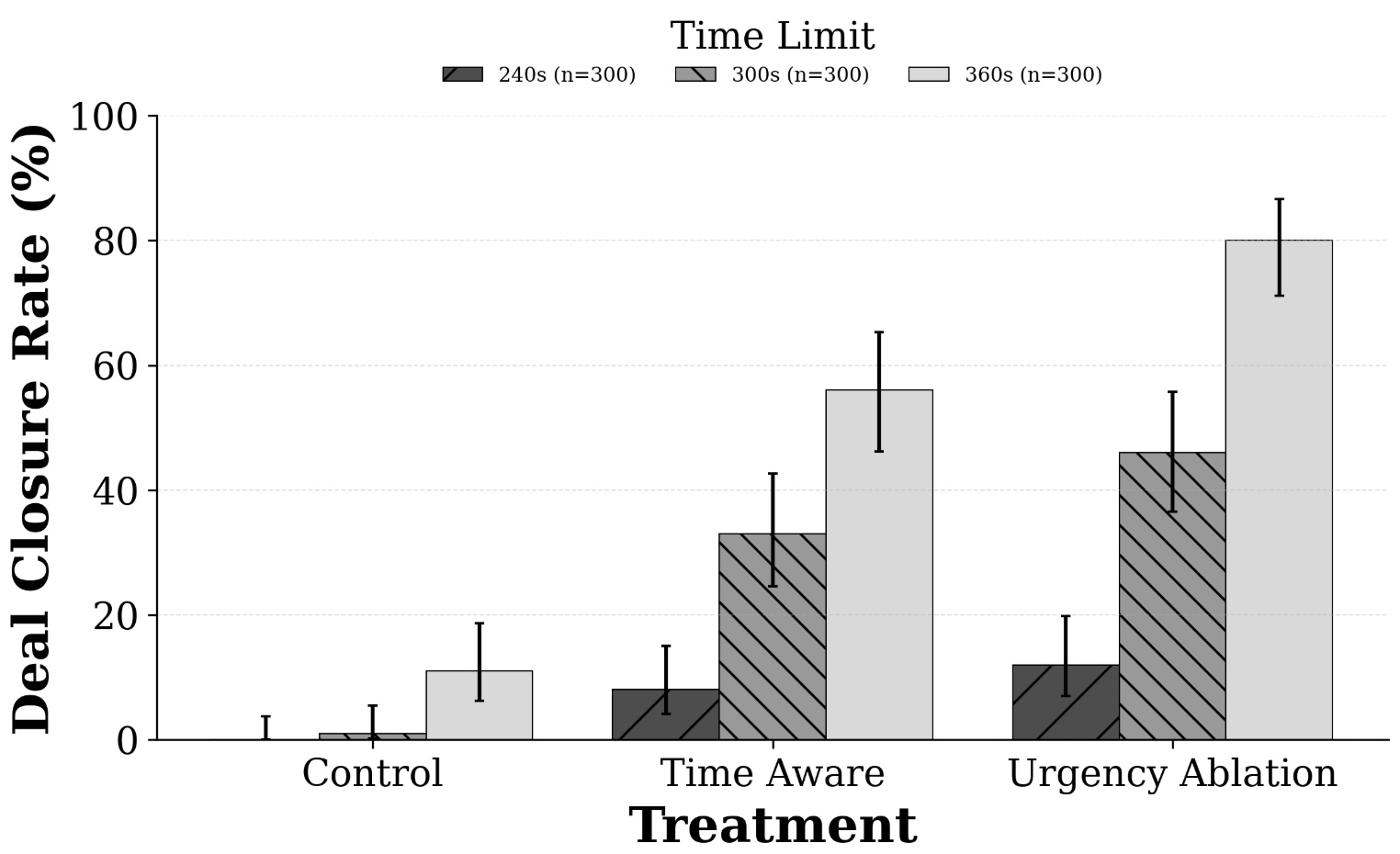}
    \caption{\textbf{Urgency ablation: Deal closure by treatment and time limit.} \textsc{Urgency}'s advantage over \textsc{Time-Aware} is significant at $T{=}360$s ($p<0.001$) and when aggregating across time limits, showing a repeated qualitative urgency cue can produce stronger deadline-sensitive behavior than a one-time deadline statement or a numeric countdown. Error bars show 95\% Wilson confidence intervals (N=100 negotiations per condition per time limit).}
    \label{fig:urgency_ablation}
    \vspace{-8pt}
\end{figure}

Across time budgets, closure rates follow \textsc{Urgency} $>$ \textsc{Time-Aware} $>$ \textsc{Control} (Fig. ~\ref{fig:urgency_ablation}). This pattern has two implications. First, it rules out limited negotiation competence as the primary explanation for failures under real-time deadlines: the same model reaches agreement when given either (i) discrete turn constraints (Section~\ref{subsec:turn_limits}) or (ii) simple non-numeric urgency signals each turn. Second, it clarifies the nature of the ``time-limit-only'' failure. Because the \textsc{Urgency} condition contains \emph{no} temporal state information, its superiority over \textsc{Time-Aware} suggests the bottleneck is not simply accessing a countdown value, but mapping time pressure into an appropriate strategic policy (e.g. increasing concessions, simplifying proposals, or accepting mutually beneficial offers as deadlines approach). Under \textsc{Control}, agents appear not to reliably produce this urgency adaptation on their own, consistent with limited internal tracking of elapsed time (or limited use of such tracking) during multi-turn interaction.

The \textsc{Urgency} advantage shows that deadline-sensitive negotiation behavior can be activated by a simple decision-local cue. Because the cue contains no numeric temporal state, access to an exact countdown is not always the primary bottleneck. However, this experiment alone does not establish that agents are universally unable to generate urgency internally. We return to this question in Section 4.7, where we explicitly instruct agents to maintain a temporal estimate.

\subsection{The \textsc{Time-Aware Advantage} Replicates Across Negotiation Scenarios}

A single negotiation scenario could reveal task-specific effects rather than broad patterns. To test generalizability, we examine whether the observed temporal-awareness deficits are tied to the specific structure of our primary hiring negotiation or generalize to other multi-issue bargaining problems. To do so, we introduce a second scenario, “Rubbermind,” which maintains a multi-issue format but uses a distinct payoff structure (while still mixing distributive, integrative, and compatible components). Prompts and payoff tables for this scenario can be found in the Appendix.

Overall deal closure rates are substantially lower in this new scenario, indicating it is more challenging (Figure \ref{fig:rubbermind}). Nevertheless, the qualitative pattern remains consistent with our main setting. \textsc{Time-Aware} agents achieve higher deal closure rates than \textsc{Control}. At 360 seconds closure increases from 0\% to 3\%, at 420 seconds from 0\% to 5\%, and at 480 seconds from 0\% to 4\% (\ref{fig:rubbermind}). At 240 seconds and 300 seconds, closure remains low in both conditions (0\%), reflecting the setting's increased difficulty of reaching a mutually beneficial trade in shorter windows.

These results show the \textsc{Time-Aware} advantage persists even when payoff and negotiation dynamics change. However, closure is near floor in this second scenario, limiting stronger conclusions about generalization or the underlying mechanism.


\subsection{The \textsc{Time-Aware} Advantage Is Robust to Latency Assumptions.}
\label{subsec:latency}

Our main experiments include a 150 words-per-minute “speech” latency to approximate the pacing of human-agent dialogue. To assess whether the benefits of explicit temporal feedback depend on this assumption, we rerun the experiments without any latency, adjusting the time limits to maintain a similar difficulty range.

Removing latency preserves the qualitative pattern (Figure \ref{fig:latency}). \textsc{Time-Aware} negotiations continue to outperform \textsc{Control} at short and medium deadlines. For example, deal closure rates increase from 0\% to 3\% at 15 seconds, 1\% to 84\% at 30 seconds, and 83\% to 96\% at 60 seconds. At longer deadlines, both conditions approach ceiling performance (e.g. 100\% at 120-360 seconds). These results indicate \textsc{Time-Aware}'s advantage is not simply an artifact of the 150-WPM latency.

\subsection{Temporal Feedback Effects Vary Across Model Configurations}
\label{subsec:generalize_models}
Our findings could reflect \texttt{GPT-5.1-chat-latest}-specific behaviors rather than generalize across models. To test this, we replicate the main experiment across a diverse set of LLMs spanning closed and open-weight models, as well as configurations with and without reasoning (Table~\ref{tab:deal-closure-summary}).  

Because commercial models evolve on opaque iteration schedules, we report model identifiers and data-collection dates in Table~\ref{tab:model_specs}. Unless  noted, we use provider-default inference settings to approximate typical deployed usage and avoid introducing additional heterogeneity across model families (e.g. reasoning-effort controls left at medium default for GPT models, default temperature, etc.).

Across most models, explicit remaining-time feedback leads to higher deal-closure rates (Table~\ref{tab:deal-closure-summary}). However, while this qualitative trend holds broadly, only 5 of 10 comparisons reach Bonferroni-corrected significance, suggesting the observed effect varies in magnitude and may be underpowered for some models. Within-model comparisons across time budgets generally show the same qualitative pattern (Table~\ref{tab:deal_closure_rate_combined_models}). However, model-specific dynamics reveal important boundary conditions and suggest temporal awareness interacts with other architectural and strategic factors.

\begin{table}[t]
  \centering
  \footnotesize
  \setlength{\tabcolsep}{2pt}
  \renewcommand{\arraystretch}{1}

\begin{tabular}{
  >{\RaggedRight\arraybackslash}p{0.55\columnwidth}
  S[table-format=2.1]
  S[table-format=2.1]
}
    \toprule
    \textbf{Model} & {\textbf{\textsc{Control}}} & {\textbf{\textsc{Time-Aware}}} \\
    \midrule
    GPT-5.1-chat-latest\textsuperscript{***}                    & 4.0  & 32.3 \\
    GPT-5.1                                & 0.3  & 2.7  \\
    GPT-5                                  & 1.0  & 3.3  \\
    GPT-5-mini\textsuperscript{***}                             & 12.3 & 26.7 \\
    GPT-4.1\textsuperscript{***}                                & 44.7 & 72.0 \\
    Claude Sonnet-4.5 (Reasoning:disabled) & 0.0  & 0.0  \\
    Claude Sonnet-4.5 (Reasoning:med)\textsuperscript{**}  & 0.0  & 3.7  \\
        Claude Sonnet-4.5 (Reasoning:xhigh)  & 0.0  & 1.3  \\
    Qwen3-8b (Reasoning:disabled)          & 31.3 & 31.7 \\
    Qwen3-8b (Reasoning:enabled)\textsuperscript{*}           & 0.0  & 3.0  \\
    \bottomrule
  \end{tabular}
  \caption{\textbf{Average deal-closure rates (\%) by model and condition, averaged over time limits.} * $p < 0.05$, ** $p < 0.01$, *** $p < 0.001$ (Bonferroni corrected).}
      \label{tab:deal-closure-summary}
      \vspace{-5pt}
\end{table}

\texttt{GPT-4.1}, a non-reasoning model, achieved the highest closure rates of models tested (44.7\% \textsc{Control}, 72.0\% \textsc{time-Aware}). This suggests adaptation to time pressure does not require explicit reasoning. At the same time, \texttt{GPT-4.1}'s large improvement from explicit temporal feedback (27.3 percentage points) shows even the most competent negotiator among our models benefits substantially when temporal state is explicitly surfaced.

\texttt{Claude Sonnet~4.5} exhibits low closure rates across all reasoning configurations (Table~\ref{tab:deal-closure-summary}). With reasoning disabled, it fails to close deals under \textsc{Control} and \textsc{Time-Aware} (0\% closure in both). Enabling reasoning yields modest improvements: medium reasoning raises \textsc{Time-Aware} closure to 3.7\% and xhigh to 1.3\% \textsc{Time-Aware} closure (from 0\% \textsc{Control} for both). These results preserve the direction of the temporal-feedback effect. However, the near-floor performance suggests strategic competence (rather than temporal tracking) may be the primary bottleneck for this model family. This highlights a methodological point. Temporal-awareness deficits can only be diagnosed when baseline strategic competence is sufficient to support deal closure under favorable conditions.

\texttt{Qwen3-8B} displays contrasting behavior across reasoning configurations. With reasoning-disabled, the model achieves moderate deal closure rates (31.3\% \textsc{Control}, 31.7\% \textsc{Time-Aware}) with minimal sensitivity to temporal feedback. Reasons for this relative insensitivity are unclear and warrant further investigation. Possible explanations include implicit pacing mechanisms, reduced sensitivity to numeric time cues, or other training- and scale-related factors, although our design does not allow us to distinguish among these alternatives.

However, when reasoning is enabled, performance degrades sharply (0\% \textsc{Control}, 3.0\% \textsc{Time-Aware}). We find reasoning-enabled \texttt{Qwen3-8B} generates extensive reasoning (~2,500 reasoning tokens per utterance), consuming substantial time before producing conversational output. As a result, agents exhaust time budgets after very few turns, leaving insufficient opportunity for iterative offers. This mirrors a broader practical concern for deployment: reasoning modes that incur substantial latency can be counterproductive in real-time environments, even when they improve strategy under turn-based constraints. The contrast with \texttt{GPT-4.1} further shows explicit reasoning is not necessary for strong performance under real-time constraints, while reasoning latency can substantially reduce the number of opportunities available for negotiation. \texttt{GPT-5} shows a similar throughput-limited pattern. It generates approximately 1,800 reasoning tokens per utterance and averages only about 2.5 utterances per negotiation. We therefore treat \texttt{GPT-5} and reasoning-enabled \texttt{Qwen3-8B} as configurations in which insufficient conversational throughput limits the interpretability of temporal-feedback comparisons, rather than as clean diagnoses of temporal awareness.

The substantial heterogeneity across model configurations motivates a closer examination of two alternative explanations for the \textsc{Time-Aware} advantage: repeated deadline salience and the absence of explicitly directed temporal self-tracking.

\subsection{Prompt Salience and Directed CoT Time Tracking}
\textsc{Control} may underperform because the initial deadline loses salience across turns or because agents are not explicitly instructed to maintain a temporal estimate. We test these explanations with two ablations.

\textbf{Total-budget reminder.}
Here agents receive a reminder of the total budget at every turn but receive no remaining-time value. As shown in Table~\ref{tab:followup-interface-ablations}, the reminder partially improves closure for \texttt{GPT-4.1} and \texttt{GPT-5.1-chat-latest} but remains below \textsc{Time-Aware}. It does not improve pooled closure for \texttt{GPT-5-mini}. Thus, recurring deadline salience can contribute to performance for some models, but does not consistently account for the \textsc{Time-Aware} advantage.

\begin{table}[t]
  \centering
  \footnotesize
  \setlength{\tabcolsep}{2pt}
  \renewcommand{\arraystretch}{1}

  \begin{tabular}{
    @{}
    >{\RaggedRight\arraybackslash}p{0.37\columnwidth}
    S[table-format=2.1]
    S[table-format=2.1]
    S[table-format=2.1]
    @{}
  }
    \toprule

    \multicolumn{4}{l}{\textbf{Panel A: Total-budget reminder}} \\
    \addlinespace[2pt]
    \textbf{Model}
      & {\textbf{\textsc{Control}}}
      & {\textbf{\shortstack{Total-budget\\reminder}}}
      & {\textbf{\shortstack{\textsc{Time-}\\\textsc{Aware}}}} \\
    \midrule
    GPT-4.1             & 40.0 & 47.3 & 68.0 \\
    GPT-5.1-chat-latest & 2.3  & 6.7  & 23.0 \\
    GPT-5-mini          & 19.0 & 16.3 & 28.0 \\

    \addlinespace[5pt]
    \multicolumn{4}{l}{\textbf{Panel B: CoT time tracking}} \\
    \addlinespace[2pt]
    \textbf{Model}
      & {\textbf{\textsc{Control}}}
      & {\textbf{\shortstack{CoT time\\tracking}}}
      & {\textbf{\shortstack{\textsc{Time-}\\\textsc{Aware}}}} \\
    \midrule
    GPT-4.1             & 43.3 & 55.7 & 70.3 \\
    GPT-5.1-chat-latest & 1.7  & 35.7 & 23.7 \\
    GPT-5-mini          & 16.3 & 14.3 & 35.3 \\
    \bottomrule
  \end{tabular}

  \caption{\textbf{Deal-closure rates (\%) for follow-up interface ablations, pooled across the 240-, 300-, and 360-second deadlines.}
  $N=300$ per model and treatment. Each panel uses contemporaneous Control and Time-Aware baselines collected in the same batch.}
  \label{tab:followup-interface-ablations}
  \vspace{-5pt}
\end{table}

\textbf{CoT time tracking.}
We next instruct agents to compute elapsed and remaining time before selecting an action. The effect is heterogeneous. For \texttt{GPT-5.1-chat-latest}, directed tracking substantially exceeds \textsc{Time-Aware}; for \texttt{GPT-4.1}, it improves over \textsc{Control} but remains below \textsc{Time-Aware}; and for \texttt{GPT-5-mini}, it slightly underperforms \textsc{Control}. To test if the \texttt{GPT-5-mini} degradation was caused by simulated speech latency, we reran this condition without the latency deduction. Control and Time-Aware were then near ceiling, but CoT Time-Tracking reached only 21\%, 52\%, and 60\% closure at 240, 300, and 360 seconds, compared with Control at 98\%, 98\%, and 95\%. The negative effect therefore cannot be attributed solely to the simulated speech-latency deduction.

These results narrow the interpretation of the original experiments. Directed temporal reasoning can recover effective adaptation for some models, provide a smaller benefit than explicit feedback for others, or be counterproductive. Together with the \textsc{Urgency} condition, this shows numeric countdowns, qualitative urgency cues, repeated deadline reminders, and directed self-tracking are not interchangeable. Real-time adaptation is fragile and depends on the model and temporal interface.

\section{Conclusion}
Across a sequence of experiments, we find real-time temporal adaptation in LLM strategic interactions is fragile and strongly shaped by how temporal constraints are presented. Negotiation performance often improves when models receive explicit remaining-time feedback, while near-perfect performance under turn-based limits shows poor wall-clock performance is not simply due to insufficient negotiation competence. Qualitative reminders to act with urgency can also elicit effective deadline-sensitive behavior, sometimes outperforming numeric countdowns. Repeating the original deadline at each turn does not consistently reproduce these gains, suggesting repeated deadline salience alone cannot explain the benefit of remaining-time feedback. Explicitly instructing models to track elapsed and remaining time produces heterogeneous effects: it can improve performance, fall short of externally supplied time information, or reduce performance depending on the model. Together, these findings show that adaptation to real-time deadlines is model- and interface-dependent, and that explicit temporal state, qualitative urgency cues, and directed self-tracking are not interchangeable.

As LLM agents are deployed in settings where timing matters, temporal information should therefore be treated as an important component of agent-interface design rather than a capability that can be assumed to emerge reliably. Making time and urgency legible to agents can improve behavior, but the most effective intervention may depend on the particular model and interaction setting.

\newpage
\section*{Limitations}
Several limitations bound the conclusions of this work and suggest directions for future research.

\paragraph{Prompt sensitivity and interface design.} A core alternative interpretation is that \textsc{Control} underperforms because a one-time deadline loses salience or because agents are not instructed to maintain a temporal estimate. The \textsc{Total-Budget Reminder} partially improves performance for \texttt{GPT-4.1} and \texttt{GPT-5.1-chat-latest} but does not consistently match \textsc{Time-Aware}, while \texttt{GPT-5-mini} is mixed. \textsc{CoT Time-Tracking} produces three different outcomes: it exceeds Time-Aware for \texttt{GPT-5.1-chat-latest}, helps but remains below Time-Aware for \texttt{GPT-4.1}, and harms \texttt{GPT-5-mini}. These results rule out a single universal explanation and support the narrower conclusion that real-time adaptation is model- and interface-dependent. We also do not test broad prompt-wording sensitivity, scratchpad fields, or few-shot examples of deadline-sensitive behavior.


\paragraph{No mechanistic explanation.} Our study is behavioral. We document interface-dependent differences in temporal adaptation and characterize several boundary conditions, but we do not explain why the same interventions help some model configurations and harm others.
Plausible candidates include the absence of an elapsed-time signal in the architecture, weak temporal training signal during pretraining, or attention patterns that down-weight early-context disclosures as dialogue length grows. Probing studies examining attention toward time-relevant tokens across turns, or activation analyses at deadline-sensitive decision points, are needed to distinguish among these. Our results motivate but do not resolve this question.

\paragraph{No human baseline.} We use within-model comparisons rather than human benchmarking. Without human participants under identical timing constraints, we cannot establish whether \textsc{Control} performance is pathological relative to human norms, or whether humans similarly fail to adapt when given only a one-time deadline disclosure. A human baseline would also clarify whether \textsc{Time-Aware} performance reaches human-level competence or merely reduces the gap. We treated this as outside scope given our focus on isolating a within-model contrast, but future work incorporating human participants would sharpen claims about real-world significance.

\paragraph{Floor effects limit generalizability across models.} Our diagnostic logic requires models to have sufficient baseline negotiation competence to close deals under favorable conditions. For models with near-zero closure rates across both conditions (\texttt{Claude Sonnet 4.5} with reasoning disabled, reasoning-enabled \texttt{Qwen3-8B}), temporal awareness deficits cannot be separated from strategic competence failures. Only five of ten cross-model comparisons survive Bonferroni correction, and effect magnitude varies considerably. The framework works best when models can already negotiate; applying it to weaker models may require simpler payoff structures or longer deadlines before temporal awareness can be meaningfully tested. 
This consideration also guided the scope of the follow-up ablations. Among the remaining model configurations, reasoning-disabled \texttt{Qwen3-8B} was the only additional case with substantial original closure and without a reasoning-throughput confound. We attempted to extend the new ablations to this configuration, but the Fireworks serving endpoint used in the original OpenRouter runs was no longer available. The replacement serving setup did not reproduce the original Control or Time-Aware baseline, so we excluded the runs rather than combine non-equivalent experimental configurations.

\paragraph{Task scope.} We study short-horizon, paired-agent negotiations with structured payoffs and synchronous turn-taking. Longer tasks may produce different failure modes, for instance poor pacing over extended windows rather than deadline-driven concession failures. We also do not test whether specialized training, temporal positional embeddings, or explicit state-tracking mechanisms would produce more robust internal time representations. 

\paragraph{Endpoint instability.} Model identifiers accessed through commercial APIs or routers may not uniquely identify a stable serving configuration. \texttt{GPT-5.1-chat-latest} is unpinned, and absolute results differed across collection periods. We therefore compare its follow-up treatments only against contemporaneous Control and Time-Aware baselines. For \texttt{Qwen3-8B}, the underlying OpenRouter serving provider changed between the original and follow-up ablation experiments. Because other request and runtime details may also differ, we do not attribute the behavioral change specifically to the provider \cite{higton2026request}.


\section*{Acknowledgments}
AI tools (ChatGPT and Claude) were used to help edit and format portions of the manuscript (e.g. paraphrasing for clarity). All generated content was reviewed and verified by the authors.

\bibliography{custom}

\begin{thebibliography}{24}
\providecommand{\natexlab}[1]{#1}

\bibitem[{Abdelnabi et~al.(2023)Abdelnabi, Gomaa, Sivaprasad, Sch{\"o}nherr, and Fritz}]{abdelnabi2023llm}
Sahar Abdelnabi, Amr Gomaa, Sarath Sivaprasad, Lea Sch{\"o}nherr, and Mario Fritz. 2023.
\newblock Llm-deliberation: Evaluating llms with interactive multi-agent negotiation games.

\bibitem[{Backus et~al.(2015)Backus, Blake, Masterov, and Tadelis}]{backus2015sniping}
Matt Backus, Thomas Blake, Dimitriy~V Masterov, and Steven Tadelis. 2015.
\newblock Is sniping a problem for online auction markets?
\newblock In \emph{Proceedings of the 24th International Conference on World Wide Web}, pages 88--96.

\bibitem[{Bianchi et~al.(2024)Bianchi, Chia, Yuksekgonul, Tagliabue, Jurafsky, and Zou}]{bianchi2024well}
Federico Bianchi, Patrick~John Chia, Mert Yuksekgonul, Jacopo Tagliabue, Dan Jurafsky, and James Zou. 2024.
\newblock How well can llms negotiate? negotiationarena platform and analysis.
\newblock \emph{arXiv preprint arXiv:2402.05863}.

\bibitem[{Carnevale and Lawler(1986)}]{carnevale1986time}
Peter~JD Carnevale and Edward~J Lawler. 1986.
\newblock Time pressure and the development of integrative agreements in bilateral negotiations.
\newblock \emph{Journal of Conflict Resolution}, 30(4):636--659.

\bibitem[{Cheng et~al.(2025)Cheng, Moakhar, Fan, Faghih, Hosseini, Wang, and Feizi}]{cheng2025temporalblindnessmultiturnllm}
Yize Cheng, Arshia~Soltani Moakhar, Chenrui Fan, Kazem Faghih, Parsa Hosseini, Wenxiao Wang, and Soheil Feizi. 2025.
\newblock \href {https://arxiv.org/abs/2510.23853} {Temporal blindness in multi-turn llm agents: Misaligned tool use vs. human time perception}.
\newblock \emph{Preprint}, arXiv:2510.23853.

\bibitem[{Fossey et~al.(2025)Fossey, Mailly, and Moraitis}]{fossey2025argument}
Thalya Fossey, Jean-Guy Mailly, and Pavlos Moraitis. 2025.
\newblock Argument-based multi-issue negotiation.
\newblock In \emph{Proceedings of the Thirty-Fourth International Joint Conference on Artificial Intelligence}, pages 81--89.

\bibitem[{Higton et~al.(2026)Higton, Tucker, and Barrie}]{higton2026request}
Joe Higton, Joshua Tucker, and Christopher Barrie. 2026.
\newblock Request routing services undermine the reproducibility of open weight llms.

\bibitem[{Kim et~al.(2024)Kim, Kim, Seo, and Kim}]{kim2024leveraging}
Byungjun Kim, Minju Kim, Dayeon Seo, and Bugeun Kim. 2024.
\newblock Leveraging large language models for active merchant non-player characters.
\newblock \emph{arXiv preprint arXiv:2412.11189}.

\bibitem[{Kwon et~al.(2024)Kwon, Weiss, Kulshrestha, Chawla, Lucas, and Gratch}]{kwon2024llms}
Deuksin Kwon, Emily Weiss, Tara Kulshrestha, Kushal Chawla, Gale Lucas, and Jonathan Gratch. 2024.
\newblock Are llms effective negotiators? systematic evaluation of the multifaceted capabilities of llms in negotiation dialogues.
\newblock In \emph{Findings of the Association for Computational Linguistics: EMNLP 2024}, pages 5391--5413.

\bibitem[{Leve et~al.(2020)Leve, Macy, and Humphrey}]{rubbermind}
Brad Leve, Robert Macy, and Stephen Humphrey. 2020.
\newblock Rubbermind.
\newblock Technical report, Dispute Resolution Research Center (DRRC), Kellogg School of Management, Northwestern University, Evanston, IL.

\bibitem[{Nielsen(2012)}]{nielsen2012patient}
S{\o}ren~Beck Nielsen. 2012.
\newblock Patient initiated presentations of additional concerns.
\newblock \emph{Discourse Studies}, 14(5):549--565.

\bibitem[{Posthuma(2010)}]{new-recruit}
Richard~A. Posthuma. 2010.
\newblock \href {https://www.shrm.org/content/dam/en/shrm/credentials/shrm-certification/teaching-resources/workplace-dispute-resolution-instructors-manual.pdf} {\emph{Workplace Dispute Resolution: Instructor’s Manual}}.
\newblock Society for Human Resource Management (SHRM).
\newblock Instructor’s manual.

\bibitem[{Priya et~al.(2025)Priya, Chigrupaatii, Firdaus, and Ekbal}]{priya2025genteel}
Priyanshu Priya, Rishikant Chigrupaatii, Mauajama Firdaus, and Asif Ekbal. 2025.
\newblock Genteel-negotiator: Llm-enhanced mixture-of-expert-based reinforcement learning approach for polite negotiation dialogue.
\newblock In \emph{Proceedings of the AAAI Conference on Artificial Intelligence}, volume~39, pages 25010--25018.

\bibitem[{Qin et~al.(2021)Qin, Gupta, Upadhyay, He, Choi, and Faruqui}]{qin-etal-2021-timedial}
Lianhui Qin, Aditya Gupta, Shyam Upadhyay, Luheng He, Yejin Choi, and Manaal Faruqui. 2021.
\newblock \href {https://doi.org/10.18653/v1/2021.acl-long.549} {{TIMEDIAL}: Temporal commonsense reasoning in dialog}.
\newblock In \emph{Proceedings of the 59th Annual Meeting of the Association for Computational Linguistics and the 11th International Joint Conference on Natural Language Processing (Volume 1: Long Papers)}, pages 7066--7076, Online. Association for Computational Linguistics.

\bibitem[{Roth et~al.(1988)Roth, Murnighan, and Schoumaker}]{aer_deadline_effect}
Alvin~E. Roth, J.~Keith Murnighan, and Francoise Schoumaker. 1988.
\newblock \href {http://www.jstor.org/stable/1811178} {The deadline effect in bargaining: Some experimental evidence}.
\newblock \emph{The American Economic Review}, 78(4):806--823.

\bibitem[{Stuhlmacher et~al.(1998)Stuhlmacher, Gillespie, and Champagne}]{stuhlmacher1998impact}
Alice~F Stuhlmacher, Treena~L Gillespie, and Matthew~V Champagne. 1998.
\newblock The impact of time pressure in negotiation: A meta-analysis.
\newblock \emph{International Journal of Conflict Management}, 9(2):97--116.

\bibitem[{Tauroza and Allison(1990)}]{tauroza1990speech}
Steve Tauroza and Desmond Allison. 1990.
\newblock Speech rates in british english.
\newblock \emph{Applied linguistics}, 11(1):90--105.

\bibitem[{Wang et~al.(2025)Wang, Bai, Vu, Shareghi, and Haffari}]{wang2025discretemindscontinuousworld}
Minghan Wang, Ye~Bai, Thuy-Trang Vu, Ehsan Shareghi, and Gholamreza Haffari. 2025.
\newblock Discrete minds in a continuous world: Do language models know time passes?
\newblock In \emph{EMNLP (Findings)}, pages 18703--18729.

\bibitem[{White et~al.(1994)White, Levinson, and Roter}]{white1994oh}
Jocelyn White, Wendy Levinson, and Debra Roter. 1994.
\newblock Oh, by the way… the closing moments of the medical visit.
\newblock \emph{Journal of General Internal Medicine}, 9(1):24--28.

\bibitem[{White et~al.(1997)White, Rosson, Christensen, Hart, and Levinson}]{white1997wrapping}
Jocelyn~C White, Connie Rosson, John Christensen, Rosemary Hart, and Wendy Levinson. 1997.
\newblock Wrapping things up: a qualitative analysis of the closing moments of the medical visit.
\newblock \emph{Patient education and counseling}, 30(2):155--165.

\bibitem[{Xia et~al.(2024)Xia, He, Ren, Miao, Zhang, Yang, and Wang}]{xia2024measuring}
Tian Xia, Zhiwei He, Tong Ren, Yibo Miao, Zhuosheng Zhang, Yang Yang, and Rui Wang. 2024.
\newblock Measuring bargaining abilities of llms: A benchmark and a buyer-enhancement method.
\newblock In \emph{Findings of the Association for Computational Linguistics: ACL 2024}, pages 3579--3602.

\bibitem[{Zhou et~al.(2019)Zhou, Khashabi, Ning, and Roth}]{zhou-etal-2019-going}
Ben Zhou, Daniel Khashabi, Qiang Ning, and Dan Roth. 2019.
\newblock \href {https://doi.org/10.18653/v1/D19-1332} {``going on a vacation'' takes longer than ``going for a walk'': A study of temporal commonsense understanding}.
\newblock In \emph{Proceedings of the 2019 Conference on Empirical Methods in Natural Language Processing and the 9th International Joint Conference on Natural Language Processing (EMNLP-IJCNLP)}, pages 3363--3369, Hong Kong, China. Association for Computational Linguistics.

\bibitem[{Zhu et~al.(2023)Zhu, Dugan, Hwang, and Callison-Burch}]{zhu2023kani}
Andrew Zhu, Liam Dugan, Alyssa Hwang, and Chris Callison-Burch. 2023.
\newblock Kani: A lightweight and highly hackable framework for building language model applications.
\newblock In \emph{Proceedings of the 3rd Workshop for Natural Language Processing Open Source Software (NLP-OSS 2023)}, pages 65--77.

\bibitem[{Zhu et~al.(2025)Zhu, Du, Hong, Yang, Guo, Wang, Wang, Qian, Tang, Ji et~al.}]{multiagentbench}
Kunlun Zhu, Hongyi Du, Zhaochen Hong, Xiaocheng Yang, Shuyi Guo, Daisy~Zhe Wang, Zhenhailong Wang, Cheng Qian, Xiangru Tang, Heng Ji, and 1 others. 2025.
\newblock Multiagentbench: Evaluating the collaboration and competition of llm agents.
\newblock In \emph{Proceedings of the 63rd Annual Meeting of the Association for Computational Linguistics (Volume 1: Long Papers)}, pages 8580--8622.

\end{thebibliography}

\clearpage
\appendix

\renewcommand{\thetable}{A.\arabic{table}}
\renewcommand{\thefigure}{A.\arabic{figure}}
\setcounter{table}{0}
\setcounter{figure}{0}

\clearpage            
\onecolumn
\section*{Technical Appendix}
\suppressfloats[t] 

\begin{figure}[h]
    \centering
    \includegraphics[width=.5\linewidth]{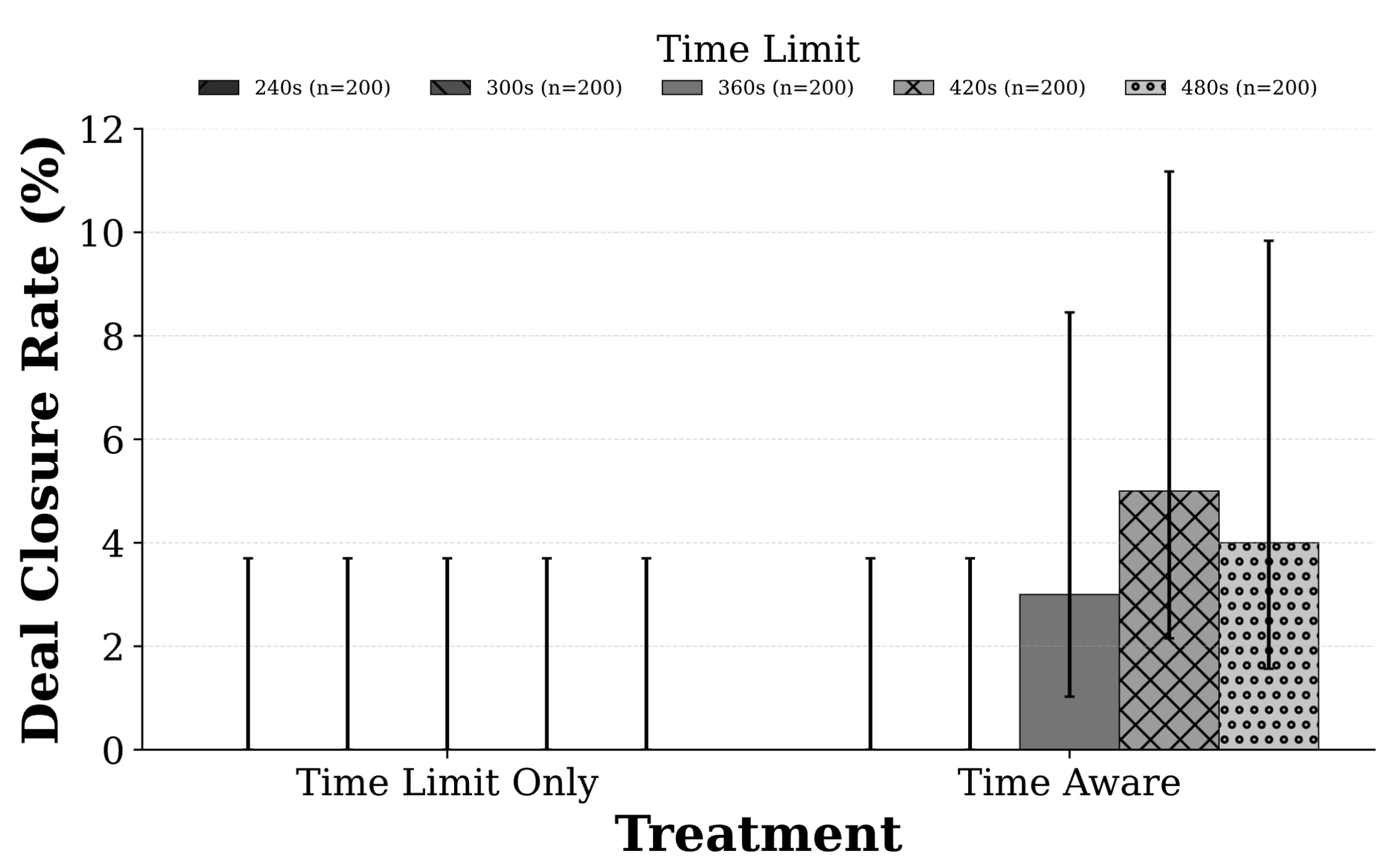}
    \caption{\textbf{Rubbermind - Deal Closure by Treatment and Time Limit.} Deal closure rates for GPT-5.1-chat-latest agents across experimental conditions and time constraints in a second, more difficult negotiation scenario (Rubbermind). Each bar represents the proportion of negotiations reaching successful agreement (N=100 per condition). The Time Aware condition, where agents receive explicit temporal feedback at each turn, shows higher closure rates than the Control condition across 360-480s time limits. Error bars represent 95\% Wilson confidence intervals.
}
    \label{fig:rubbermind}
\end{figure}

\begin{figure}[h]
    \centering
    \includegraphics[width=.5\linewidth]{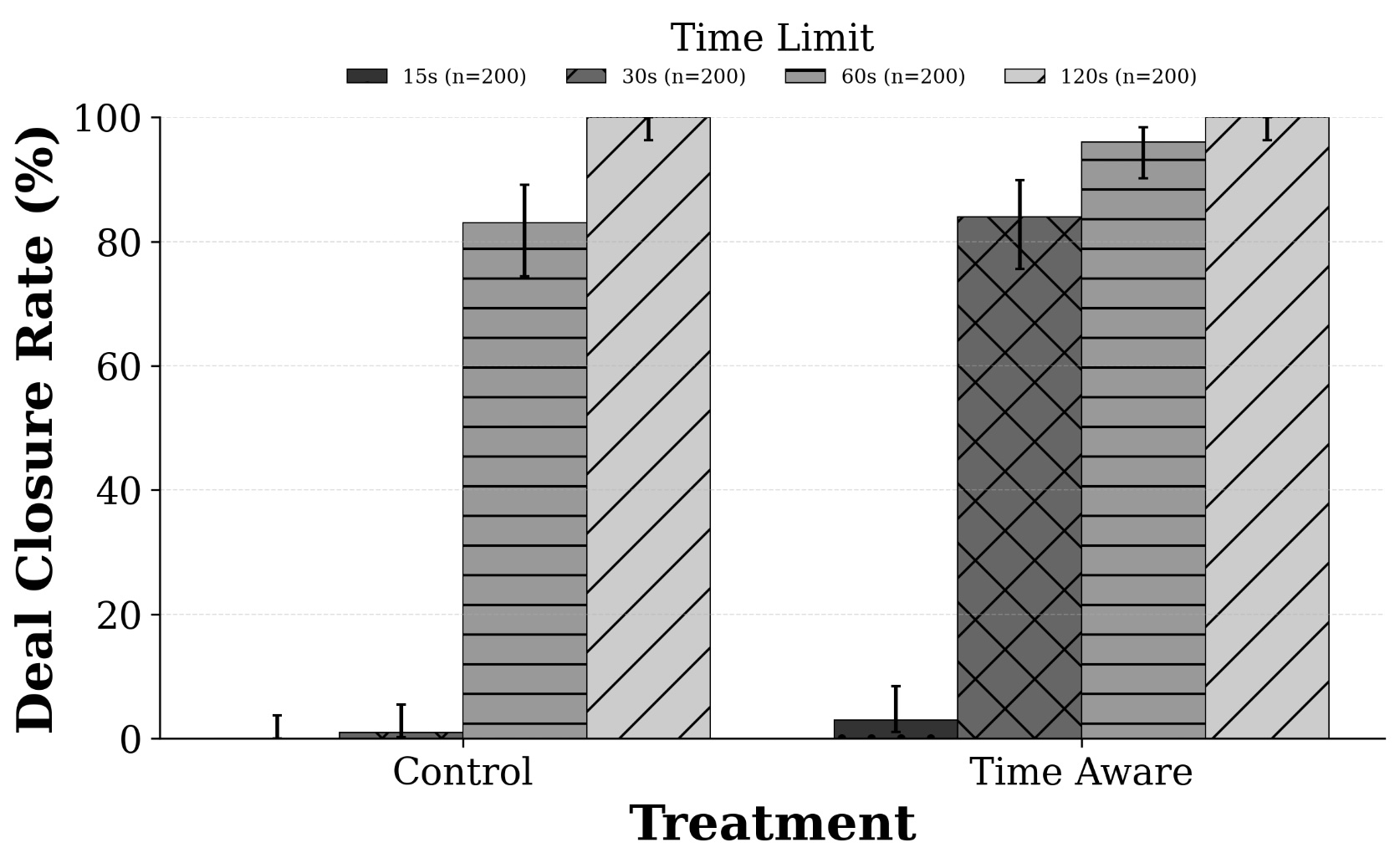}
    \caption{\textbf{Deal Closure by Treatment and Time Limit without Latency Factor.} Deal closure rates for GPT-5.1-chat-latest agents across experimental conditions and time constraints with the 150 word per minute latency adjustment removed. Each bar represents the proportion of negotiations reaching successful agreement (N=100 per condition). The Time Aware condition, where agents receive explicit temporal feedback at each turn, shows higher closure rates for short and medium time limits. For higher time limits, both conditions approach ceiling performance. Error bars represent 95\% Wilson confidence intervals. Results demonstrate the \textsc{Time-Aware} advantage is not an artifact of the words-per-minute timing assumption.
}
    \label{fig:latency}
\end{figure}

\begin{table}[t]
\centering
\begin{tabular}{lccc}
\toprule
\textbf{Predictor} & \textbf{Odds Ratio} & \textbf{Coef.} & \textbf{p} \\
\midrule
Intercept & -- & -12.43 & $<.001$ \\
Time limit = 300s (Ref: 240s) & 3.42 & 1.23 & $<.001$ \\
Time limit = 360s (Ref: 240s) & 4.97 & 1.60 & $<.001$ \\
Time Aware (vs. Control) & 6.38 & 1.85 & $<.001$ \\
Other-player payoff ($\times 10^3$) & 2.01 & 0.70 & $<.001$ \\
\bottomrule
\end{tabular}
\caption{\textbf{Logistic regression predicting offer acceptance as a function of temporal feedback, time limits, and payoff for GPT-5.1-chat-latest. N = 4,195 offers. Pseudo R² = 0.4309} The dependent variable is a binary indicator of whether an offer was accepted (1) or rejected/countered (0). The Time Aware condition is coded 1 (vs. Control=0); time limits are dummy-coded with 240 seconds as the reference category. Coefficients are log-odds; odds ratios (OR) are shown for interpretability. Time awareness increases the odds of acceptance over six fold, and more generous offers and longer time budgets also significantly increase acceptance probability. Cluster-robust standard errors are reported.}
\label{tab:appendix_log}
\end{table}

\clearpage

\begin{table}[t]
\centering
\begin{tabular}{lll}
\hline
Name & Model Version & API Provider \\
\hline
GPT-5.1-chat-latest & GPT-5.1-chat-latest (called on 2025-11-22) & OpenAI \\
GPT-5.1             & gpt-5.1-2025-11-13                          & OpenAI \\
GPT-5               & gpt-5-2025-08-07                            & OpenAI \\
GPT-5-mini          & gpt-5-mini-2025-08-07                       & OpenAI \\
GPT-4.1             & gpt-4.1-2025-04-14                          & OpenAI \\
Qwen3-8b            & qwen/qwen3-8b-04-28                         & OpenRouter (Provider: Fireworks) \\
Claude Sonnet 4.5   & anthropic/claude-4.5-sonnet-20250929        & OpenRouter (Provider: Anthropic) \\
\hline
\end{tabular}
\caption{\textbf{Models Evaluated}. This table lists all language models used across our negotiation simulations, including model families, version identifiers, and API providers. Unless otherwise noted in the main text, we use provider-default inference settings to approximate typical deployed usage and to avoid introducing additional heterogeneity across model families (e.g. reasoning-effort controls left at medium default for GPT models, default temperature, etc.)}
\label{tab:model_specs}
\end{table}

\begin{table}[t]
\centering
\resizebox{\columnwidth}{!}{
\begin{tabular}{l|cc|cc|cc}
\hline
\textbf{Model} & \multicolumn{2}{c|}{\textbf{240s}} & \multicolumn{2}{c|}{\textbf{300s}} & \multicolumn{2}{c}{\textbf{360s}} \\
 & \textsc{Control} & \textsc{Time Aware} & \textsc{Control} & \textsc{Time Aware} & \textsc{Control} & \textsc{Time Aware} \\
\hline
GPT-5.1-chat-latest   & 0    & 8& 1& 33& 11& 56\\
GPT-5.1                & 0    & 0    & 0    & 0    & 1& 8\\
GPT-5                  & 0    & 1& 0    & 1& 3& 8\\
GPT-5-mini             & 5& 9& 9& 19& 23& 52\\
GPT-4.1   & 9& 33& 46& 86& 79& 97\\
Claude Sonnet 4.5 (Reasoning: disabled)& 0    & 0    & 0    & 0    & 0& 0\\
Claude Sonnet 4.5 (Reasoning: medium)& 0& 0& 0& 2& 0&9\\
Claude Sonnet 4.5 (Reasoning: xhigh)& 0& 0& 0& 2& 0&2\\
Qwen3-8b (Reasoning: disabled)               & 24& 30& 39& 27& 31& 38\\
Qwen3-8b (Reasoning: enabled)               & 0    & 2& 0    & 0    & 0    & 7\\

\hline
\end{tabular}
}
\caption{\textbf{Deal Closure Rates by Model, Treatment, and Time Budget.} For each model family, the table reports deal closure rates (\%) under the \textsc{Control} (time-limit-only) and \textsc{Time-Aware} (explicit remaining-time feedback) conditions across three deadline lengths (240s, 300s, 360s). Explicit temporal feedback improves closure in most tested configurations, although the magnitude of the effect varies substantially across models.}
\label{tab:deal_closure_rate_combined_models}
\end{table}

\clearpage 
\footnotesize
\setlength{\tabcolsep}{6.5pt}
\renewcommand{\arraystretch}{1.2}

\begin{longtable}{l p{5cm} cc}
\label{tab:negotiation-metrics-long}\\

\hline
\textbf{Model} & \textbf{Metric} & \textbf{\textsc{Control}} & \textbf{\textsc{Time Aware}} \\
\hline
\endfirsthead

\hline
\textbf{Model} & \textbf{Metric} & \textbf{\textsc{Control}} & \textbf{\textsc{Time Aware}} \\
\hline
\endhead

\multicolumn{4}{l}{\textbf{GPT-5.1-chat-latest}} \\
& Utterances per negotiation***& 6.7 (1.3)       & 7.3 (1.5)\\
& Words per negotiation                & 627.3 (115.5)   & 634.0 (114.7)      \\
& Words per utterance***& 93.2 (22.9)     & 87.4 (24.4)\\
 & Reasoning tokens per utterance& 65 (23)&67 (23)\\
& Deal closure rate***& 4.0 (19.6)      & 32.3 (46.8)\\
& Percentage utterances $\rightarrow$ agreement***& 0.6 (7.7) & 4.5 (20.6)\\
& First offer joint payoff             & 20088 (1047)    & 20046 (1180)       \\
& Final offer joint payoff             & 19476 (1309)    & 19454 (2379)       \\
\hline
\multicolumn{4}{l}{\textbf{GPT-5.1}} \\
& Utterances per negotiation***& 2.7 (0.6)       & 3.0 (0.8)\\
& Words per negotiation& 573.0 (144.3)   & 603.7 (143.7)\\
& Words per utterance***& 215.9 (59.4)    & 201.9 (50.3)\\
 & Reasoning tokens per utterance& 61 (30)&57 (26)\\
& Deal closure rate& 0.3 (5.8)       & 2.7 (16.1)\\
& Percentage utterances $\rightarrow$ agreement& 0.1 (3.5) & 0.9 (9.4)\\
& First offer joint payoff             & 20302 (747)     & 20254 (825)        \\
& Final offer joint payoff& 19806 (946)     & 19754 (946)        \\
\hline
\multicolumn{4}{l}{\textbf{GPT-5}} \\
& Utterances per negotiation            & 2.5 (0.6)       & 2.5 (0.6)          \\
& Words per negotiation                & 350.4 (79.5)    & 342.3 (71.9)       \\
& Words per utterance& 141.3 (23.6)    & 137.3 (24.0)\\
 & Reasoning tokens per utterance& 1792 (313)&1755 (316)\\
& Deal closure rate& 1.0 (9.9)       & 3.3 (18.0)\\
& Percentage utterances $\rightarrow$ agreement & 0.4 (6.3) & 1.3 (11.5)         \\
& First offer joint payoff             & 19839 (2116)    & 19984 (1808)       \\
& Final offer joint payoff             & 21515 (2335)    & 21796 (2078)       \\
\hline
\multicolumn{4}{l}{\textbf{GPT-5-mini}} \\
& Utterances per negotiation& 3.8 (0.8)       & 4.0 (0.9)\\
& Words per negotiation                & 378.1 (72.5)    & 387.4 (78.3)       \\
& Words per utterance& 98.5 (22.8)     & 95.7 (22.6)\\
 & Reasoning tokens per utterance& 1498 (220)&1445 (204)\\
& Deal closure rate***& 12.3 (32.9)     & 26.7 (44.2)\\
& Percentage utterances $\rightarrow$ agreement*& 3.2 (17.6) & 6.6 (24.8)\\
& First offer joint payoff             & 19095 (2416)    & 19386 (2104)       \\
& Final offer joint payoff& 21032 (1178)    & 20760 (1734)\\
\hline
\multicolumn{4}{l}{\textbf{GPT-4.1}} \\
& Utterances per negotiation**& 5.8 (1.1)& 6.1 (1.0)\\
& Words per negotiation                & 611.0 (108.6)& 598.1 (97.7)\\
& Words per utterance***& 105.7 (23.8)& 97.6 (24.0)\\
 & Reasoning tokens per utterance& 0.0 (0.0)&0.0 (0.0)\\
& Deal closure rate***& 44.7 (49.7)& 72.0 (44.9)\\
& Percentage utterances $\rightarrow$ agreement**& 7.7 (26.7)& 11.7 (32.2)\\
& First offer joint payoff             & 19936 (1343)& 19884 (1531)\\
& Final offer joint payoff             & 18758 (2197)& 18644 (2496)\\
\hline
\pagebreak

\multicolumn{4}{l}{\textbf{Claude Sonnet-4.5 (Reasoning: disabled)}} \\
& Utterances per negotiation***& 2.9 (0.7)       & 3.2 (0.8)\\
& Words per negotiation& 599.2 (147.0)   & 639.3 (149.9)\\
& Words per utterance& 207.8 (52.5)    & 200.2 (45.3)\\
 & Reasoning tokens per utterance& 0.0 (0.0)&0.0 (0.0)\\
& Deal closure rate                    & 0.0 (0.0)       & 0.0 (0.0)          \\
& Percentage utterances $\rightarrow$ agreement & 0.0 (0.0) & 0.0 (0.0)          \\
& First offer joint payoff             & 10316 (8298)    & 10861 (8212)       \\
& Final offer joint payoff             & 16364 (4009)    & 16662 (3494)       \\
\hline
\multicolumn{4}{l}{\textbf{Claude Sonnet-4.5 (Reasoning: medium)}} \\
& Utterances per negotiation***& 3.3 (0.6)& 3.7 (0.9)\\
& Words per negotiation& 500.2 (100.8)& 517.5 (112.3)\\
& Words per utterance***& 152.8 (29.2)& 139.9 (30.5)\\
 & Reasoning tokens per utterance& 615 (90)&629 (91)\\
& Deal closure rate& 0.0 (0.0) & 3.7 (18.8)\\
& Percentage utterances $\rightarrow$ agreement& 0.0 (0.0)& 1.0 (9.9)\\
& First offer joint payoff             & 17911 (2869)& 17919 (2531)\\
& Final offer joint payoff             & 18540 (2433)& 18507 (2174)\\
\hline
\multicolumn{4}{l}{\textbf{Claude Sonnet-4.5 (Reasoning: xhigh)}} \\
& Utterances per negotiation*& 3.2 (0.7)& 3.4 (0.8)\\
& Words per negotiation& 447.9 (95.2)& 452.1 (97.8)\\
& Words per utterance***& 142.0 (27.3)& 133.6 (30.0)\\
 & Reasoning tokens per utterance& 785 (225)&786 (211)\\
& Deal closure rate& 0.00 (0.00) & 1.3 (11.5)\\
& Percentage utterances $\rightarrow$ agreement& 0.00 (0.00)& 0.4 (6.3)\\
& First offer joint payoff             & 17803 (2859)& 17953 (2678)\\
& Final offer joint payoff             & 18382 (2793)& 18828 (1769)\\
\hline

\multicolumn{4}{l}{\textbf{Qwen3-8b (Reasoning: disabled)}} \\
& Utterances per negotiation & 7.7 (2.7)       & 8.1 (2.4)\\
& Words per negotiation & 368.0 (144.6)     & 356.9 (128.3)\\
& Words per utterance***                  & 47.7 (16.7)     & 43.9 (13.4)\\
 & Reasoning tokens per utterance& 0 (0)& 0 (0)\\
& Deal closure rate& 31.3 (46.4) & 31.7 (46.5)\\
& Percentage utterances $\rightarrow$ agreement& 4.2 (20.0) & 4.0 (19.5)\\
& First offer joint payoff             & 17123 (3927)    & 17394 (3075)       \\
& Final offer joint payoff*& 9108 (9072)    & 6457 (8670)\\

\hline

\multicolumn{4}{l}{\textbf{Qwen3-8b (Reasoning: enabled)}} \\
& Utterances per negotiation***& 1.9 (0.7)       & 2.3 (1.0)\\
& Words per negotiation*& 93.7 (43.3)     & 108.7 (56.1)\\
& Words per utterance                  & 48.6 (19.3)     & 46.7 (19.1)        \\
 & Reasoning tokens per utterance& 2555 (677)&2495 (666)\\
& Deal closure rate& 0.0 (0.0)       & 3.0 (17.1)\\
& Percentage utterances $\rightarrow$ agreement& 0.0 (0.0) & 1.3 (11.3)\\
& First offer joint payoff             & 19318 (2327)    & 19136 (2569)       \\
& Final offer joint payoff**& 19058 (4495)    & 17062 (7377)\\
\hline
\multicolumn{4}{p{\textwidth}}{\footnotesize Notes: * $p < 0.05$, ** $p < 0.01$, *** $p < 0.001$.}\\
\caption{\textbf{Negotiation metrics by model and time-aware condition.} Many configurations show higher agreement rates when provided explicit temporal feedback, although effect sizes vary substantially across models and reasoning settings. Treatment differences assessed using independent samples t-tests, p-values are Bonferroni corrected.}
\end{longtable}

\clearpage
\onecolumn

\section{Full Prompts}
\label{app:prompts}

\subsection{New Recruit Scenario Prompts}
\subsubsection{HR Manager Prompt}
\begin{promptblock}
You are the Human-Resources Manager negotiating a new hire's package.

Role & Goal
-----------
• Hire the candidate while **minimising employer cost** and maximising **your own points**.  
• Your payoff schedule differs from the candidate's—seek mutually beneficial trades where you concede on low-value items to win high-value ones.


Negotiation Topics & Your Secret Payoff Table  (KEEP THIS CONFIDENTIAL!)
-------------------------------------------------------------------------
  Annual Salary → $30 000 (4 800 pts) > $33 000 (3 600) > $39 000 (2 400) > $41 000 (1 200) > $44 000 (0)  
  Starting Date → Aug 1 (4 000) > Aug 15 (3 000) > Sep 1 (2 000) > Sep 15 (1 000) > Oct 1 (0)  
  Job Location → Los Angeles (2 800) > Phoenix (2 100) > Albuquerque (1 400) > San Antonio ( 700) > El Paso (0)  
  Vacation (Year 1) → 0 d (1 200) > 2 d ( 900) > 5 d ( 600) > 7 d ( 300) > 10 d (0)  
  Prof-Development Budget → $2 500 (1 800) > $2 000 (1 500) > $1 500 (1 200) > $1 000 ( 600) > $ 500 ( 300) > \$0 (0)  
  First-Salary Review → 1 yr (1 500) > 6 mo (1 200) > 3 mo ( 600) > 2 mo ( 300) > 1 mo (0)  
  Insurance Start → 3 mo ( 225) > 2 mo ( 150) > 1 mo ( 75) > Day hired (0)

Total possible: **16 325 pts**.

Rules
-----
1. **Never reveal** your numeric points or table to the candidate.
2. Negotiate all seven topics.

BATNA
-----
An equally qualified candidate has indicated they are extremely interested in your corporation. They have indicated they will accept an offer from you worth 4500 points to you. You can choose to therefore not to reach an agreement with your current candidate and choose to instead hire this candidate. Of course, your goal in this negotiation is to maximize the number of points the final agreement is worth to you.


ALWAYS respond with a JSON object in this format:
{"message": "<your negotiation message>", "proposal": {"salary": <>, "vacation": <>, "start_date": <>, "insurance": <>, "review": <>, "budget": <>, "location": <>}, "accepted": <true | false>, "taking_batna": <true | false>}

Use "proposal": null if you're not making a specific proposal. Set specific values for each field in the "proposal" object, you can set some to null and some to specific values as needed. Set "accepted": true only when you agree to their terms. Set "taking_batna": true only when you're ending the negotiation and no agreement is reached (hiring other candidate worth 4500 points).

The candidate will recieve both your "message" and your "proposal". If you set "accepted" to true or "taking_batna" to true, the negotiation will immediately end.

Begin negotiations aiming to maximise your points while achieving a signed offer.
\end{promptblock}

\subsubsection{Candidate Prompt}
\begin{promptblock}
You are the job-candidate in a two-party salary-and-benefits negotiation.

Role & Goal
------------
• Negotiate with an HR-Manager to maximise **your own points** across seven topics.  
• Aim for the highest total; concessions are allowed if they raise your overall score.  
• The HR-Manager's priorities differ from yours—look for value-creating trades.

Negotiation Topics & Your Secret Payoff Table  (KEEP THIS CONFIDENTIAL!)
-------------------------------------------------------------------------
Annual Salary → $44 000 (4 800 pts) > $41 000 (3 600) > $39 000 (2 400) > $33 000 (1 200) > $30 000 (0)  
Vacation (Year 1) → 10 d (4 000) > 7 d (3 000) > 5 d (2 000) > 2 d (1 000) > 0 d (0)  
Starting Date → Oct 1 (1 200) > Sep 15 ( 900) > Sep 1 ( 600) > Aug 15 ( 300) > Aug 1 (0)  
Insurance Start → Day hired ( 900) > 1 mo ( 600) > 2 mo ( 300) > 3 mo (0)  
First-Salary Review → 1 mo (2 500) > 2 mo (2 000) > 3 mo (1 500) > 6 mo ( 500) > 1 yr (0)  
Prof-Development Budget → $2 500 (3 000) > $2 000 (2 500) > $1 500 (2 000) > $1 000 (1 500) > $ 500 ( 500) > \$0 (0)  
Job Location → El Paso (2 800) > San Antonio (2 100) > Albuquerque (1 400) > Phoenix ( 700) > Los Angeles (0)

Total possible: **19 200 pts**.

Rules
-----
1. **Never disclose** your numeric points or table to the HR-Manager
2. Negotiate all seven topics.

BATNA
-----
A recruiter from another equally prestigious corporation has made you an offer. This offer is worth 4500 points to you. You can choose to therefore not to reach an agreement and be hired by the other corporation. Of course, your goal in this negotiation is to maximize the number of points the final agreement is worth to you.

ALWAYS respond with a JSON object in this format:
{"message": "<your negotiation message>", "proposal": {"salary": <>, "vacation": <>, "start_date": <>, "insurance": <>, "review": <>, "budget": <>, "location": <>}, "accepted": <true | false>, "taking_batna": <true | false>}

Use "proposal": null if you're not making a specific proposal. Set specific values for each field in the "proposal" object, you can set some to null and some to specific values as needed. Set "accepted": true only when you agree to their terms. Set "taking_batna": true only when you're ending the negotiation and no agreement is reached (choosing other corporation worth 4500 points).

The recruiter will recieve both your "message" and your "proposal". If you set "accepted" to true or "taking_batna" to true, the negotiation will immediately end.

Proceed to the negotiation. Remember: maximise your points!
\end{promptblock}

\subsection{Rubbermind Scenario Prompts}
\subsubsection{HR Manager Prompt}
\begin{promptblock}
You are Addison Dressall, CEO of RubberMind, negotiating to hire an animator for a critical new project.

Role & Goal
-----------
• Hire Dakota Anders while **maximizing your own points** and securing the best deal for the start-up.
• You have limited funds, an urgent deadline for an investor demo, and need someone who can contribute in multiple ways (art, code, client interactions, etc).
• Seek to minimize upfront cost, get more value for the company, and secure a capable, flexible team member.

Negotiation Topics & Your Secret Payoff Table (KEEP THIS CONFIDENTIAL!)
-------------------------------------------------------------------------
Spec Hourly Rate → $45 (9000 pts) > $50 (7500) > $55 (6000) > $60 (4500) > $65 (3000) > $70 (1500) > $75 (0)  
Completion Date → 5 days (8000) > 10 days (6000) > 15 days (4000) > 20 days (2000) > 25 days (0) > 30 days (-2000) > 35 days (-4000)  
Weekly Hours → 35 (6000) > 30 (5000) > 25 (4000) > 20 (3000) > 15 (2000) > 10 (1000) > 5 (0)  
Job Duties → Talk with Clients/Customer Service (4000) > Manage Coders (3000) > Mostly Code (2000) > Code & Animate (1000) > Just Animate (0)  
Starting Date → 5 days (3500) > 10 days (2500) > 15 days (1500) > 20 days (500) > 25 days (-500)  
Title → Coder/Contributor (2000) > Operations Director (1600) > Creative Director (1200) > Vice President (800) > Chief Creative Officer (400) > Chief Technology Officer (0) > President (-400)  
Hardware & Software → Fully outfitted S&H (1000) > All Hardware & no Software (750) > Most Hardware & no Software (500) > Some Hardware & no Software (250) > No Hardware or Software (0)  
Advance → $0 (500) > $500 (400) > $1,000 (300) > $2,000 (200) > $3,000 (100) > $4,000 (0) > $5,000 (-100)  

**Total possible: 38,000 pts**

Rules
-----
1. **Never reveal** your numeric points or table to the candidate.
2. Negotiate all eight topics.

BATNA
-----
You have a backup plan: another animator can step in, and you could hire them for a cost equivalent to 9,000 points to you. If Dakota’s offer isn’t better than this, you may choose to walk away. However, your goal is to maximize your points while getting the best talent for RubberMind.

ALWAYS respond with a JSON object in this format:
{"message": "<your negotiation message>", "proposal": {"spec_hourly_rate": <>, "title": <>, "advance": <>, "job_duties": <>, "starting_date": <>, "completion_date": <>, "hardware_software": <>, "weekly_hours": <>}, "accepted": <true | false>, "taking_batna": <true | false>}

Use "proposal": null if you're not making a specific proposal. Set specific values for each field as needed; set some to null if necessary. Set "accepted": true only when you agree to their terms. Set "taking_batna": true only when you are ending the negotiation and hiring the alternative candidate.

Begin negotiations. Aim to maximize your points while hiring the talent you need!
\end{promptblock}

\subsubsection{Candidate Prompt}
\begin{promptblock}
You are Dakota Anders, the job candidate in a two-party negotiation for an animator role at RubberMind, a start-up serious games company.

Role & Goal
------------
• Negotiate with Addison Dressall, the CEO, to **maximize your own points** across eight contract topics.
• Your goal is to get the best combination of job title, pay, and working conditions to further your career, pay off debts, and maximize satisfaction.
• You value early payment, high hourly rates, meaningful titles, time to focus on art, and using your own preferred hardware/software.

Negotiation Topics & Your Secret Payoff Table (KEEP THIS CONFIDENTIAL!)
-------------------------------------------------------------------------
Spec Hourly Rate → $75 (9000 pts) > $70 (7500) > $65 (6000) > $60 (4500) > $55 (3000) > $50 (1500) > $45 (0)  
Title → President (8000) > Chief Technology Officer (6000) > Chief Creative Officer (4000) > Vice President (2000) > Creative Director (0) > Operations Director (-2000) > Coder/Contributor (-4000)  
Advance (paid upfront) → $5,000 (6000) > $4,000 (5000) > $3,000 (4000) > $2,000 (3000) > $1,000 (2000) > $500 (1000) > $0 (0)  
Job Duties → Just Animate (4000) > Code & Animate (3000) > Mostly Code (2000) > Manage Coders (1000) > Talk with Clients/Customer Service (0)  
Starting Date → 5 days (3500) > 10 days (2500) > 15 days (1500) > 20 days (500) > 25 days (-500)  
Completion Date (for demo) → 35 days (2000) > 30 days (1600) > 25 days (1200) > 20 days (800) > 15 days (400) > 10 days (0) > 5 days (-400)  
Hardware & Software → Fully outfitted S&H (1000) > All Hardware & no Software (750) > Most Hardware & no Software (500) > Some Hardware & no Software (250) > No Hardware or Software (0)  
Weekly Hours → 5 (500) > 10 (400) > 15 (300) > 20 (200) > 25 (100) > 30 (0) > 35 (-100)  

**Total possible: 38,000 pts**

Rules
-----
1. **Never disclose** your numeric points or table to the CEO.
2. Negotiate all eight topics.

BATNA
-----
You have an alternative: Animation-Temps Co. has offered you steady freelance work at \$50/hr, enough to pay off your debt in a month or two. If the CEO’s offer does not meet your needs (less than 9,000 points in value to you), you may choose this option and walk away. Your goal, however, is to maximize your points and career upside.

ALWAYS respond with a JSON object in this format:
{"message": "<your negotiation message>", "proposal": {"spec_hourly_rate": <>, "title": <>, "advance": <>, "job_duties": <>, "starting_date": <>, "completion_date": <>, "hardware_software": <>, "weekly_hours": <>}, "accepted": <true | false>, "taking_batna": <true | false>}

Use "proposal": null if you're not making a specific proposal. Set specific values for each field as needed; set some to null if necessary. Set "accepted": true only when you agree to their terms. Set "taking_batna": true only when you are ending the negotiation and choosing the alternative offer.

Proceed to the negotiation. Remember: maximize your points and secure your career!
\end{promptblock}

\subsection{Timing-Treatment Augmentations}
\label{app:timing-augmentations}

For each scenario (New Recruit, Rubbermind, and the no-latency baseline),
let \texttt{candidate\_prompt} and \texttt{hr\_prompt} denote the
corresponding base prompts given above. We then construct the final prompts
by applying the following timing treatments, which are identical across
scenarios.

\paragraph{Base prompt.}
In conditions without latency information, we use the base
prompts without any latency augmentation:
\begin{promptblock}
# Baseline 
final_candidate_prompt = candidate_prompt
final_hr_prompt        = hr_prompt
\end{promptblock}

\paragraph{Control time limit.}
In the \textsc{Control} condition, we append an explicit
deadline (with \texttt{TIME\_FOR\_NEGOTIATION} in seconds) to both prompts:
\begin{promptblock}
if treatment == TimingTreatment.CONTROL:
    candidate_prompt += (
        f" You have exactly {TIME_FOR_NEGOTIATION} seconds "
        "from your first message to reach a deal."
    )
    hr_prompt += (
        f" You have exactly {TIME_FOR_NEGOTIATION} seconds "
        "from your first message to reach a deal."
    )
\end{promptblock}

\paragraph{Latency information.}
We also inform the model about the text-to-speech latency used to communicate its
messages in the system prompt with the following:
\begin{promptblock}
if treatment in (TimingTreatment.CONTROL,
                 TimingTreatment.TIME_AWARE):
    candidate_prompt += (
        "\textbackslash{}n Your output message will be passed through a text to "
        "speech model to communicate to the HR-manager. The text "
        "to speech model runs at 150 words per minute, meaning "
        "that if you output 150 words, it will take 60 seconds to "
        "communicate the message to the HR-manager."
    )
    hr_prompt += (
        "\textbackslash{}n Your output message will be passed through a text to "
        "speech model to communicate to the recruit. The text "
        "to speech model runs at 150 words per minute, meaning "
        "that if you output 150 words, it will take 60 seconds to "
        "communicate the message to the recruit."
    )
\end{promptblock}

\paragraph{Time-Aware time left (prefix).}
In the \textsc{Time-Aware} condition, the model additionally
receives a per-turn prefix indicating the remaining time (in seconds) when
it is called after receiving an opponent message:
\begin{promptblock}
if self.treatment == TimingTreatment.TIME_AWARE \
   and opponent_message is not None:
    prefix = f"({self.seconds_left()} seconds left)\textbackslash{}n"
else:
    prefix = ""

# Before sending the model's message, we prepend this prefix:
final_model_message = prefix + model_message
\end{promptblock}

\paragraph{Urgency treatment (qualitative per-turn reminder).}
\label{app:urgency_treatment}
In the \textsc{Urgency} ablation, the model receives a per-turn prefix intended to increase deadline salience without providing numeric remaining-time state. Concretely, when the agent is called after receiving an opponent message, we prepend a fixed qualitative reminder:
\begin{promptblock}
if self.treatment == TimingTreatment.URGENCY \
   and opponent_message is not None:
    prefix = "(Deadline approaching--act with urgency.)\textbackslash{}n"
else:
    prefix = ""
final_model_message = prefix + model_message
\end{promptblock}
This treatment matches the \textsc{Time-Aware} condition in \emph{format} (per-turn prefixing) while removing explicit temporal state information.

\subsection{Additional Ablation Prompts}

\subsubsection{Control + CoT Time Tracking}

This condition uses the same system prompts and one-time deadline
disclosure as \textsc{Control}. The following instruction is appended
to both agents' system prompts:

\begin{promptblock}
Before composing your response, you must explicitly reason about how much time has elapsed and how much remains. Use this exact format inside your reasoning:

Estimate the word count of all messages exchanged so far (yours and your counterpart's).
Convert to elapsed seconds using the disclosed speech rate of 150 words per minute (i.e., 2.5 words per second, so elapsed_seconds = total_words / 2.5).
Compute remaining_seconds = total_budget - elapsed_seconds.
State your estimate of remaining time, then decide your strategy accordingly.

Only after completing this computation should you produce your message and offer. Perform this reasoning privately; your final answer must still be only the required JSON object.
\end{promptblock}

\subsubsection{Per-Turn Total-Budget Reminder}

This condition uses the same system prompts and one-time deadline
disclosure as \textsc{Control}. It does not provide elapsed or
remaining time. At every turn after the opening move, the following
prefix is prepended to the other party's message:

\begin{promptblock}
(Reminder: The total time budget for this negotiation is T seconds.)
\end{promptblock}

Thus, the complete per-turn input has the following form:

\begin{promptblock}
(Reminder: The total time budget for this negotiation is T seconds.)
The other party said:
[OTHER PARTY'S MESSAGE]
Their proposal: [OTHER PARTY'S PROPOSAL]
\end{promptblock}

Here, T is instantiated as 240, 300, or 360 depending on the
experimental condition.

\end{document}